%% file: template.tex
\documentclass{article}

\usepackage{arxiv}

\usepackage[utf8]{inputenc} 
\usepackage[T1]{fontenc}    
\usepackage{hyperref}       
\usepackage{url}            
\usepackage{booktabs}       
\usepackage{nicefrac}       
\usepackage{microtype}      
\usepackage{lipsum}
\usepackage{graphicx}
\usepackage{multirow}%
\usepackage{subcaption}
\usepackage{amsmath,amssymb,amsfonts}%
\usepackage{amsthm}%
\usepackage{mathrsfs}%
\usepackage[title]{appendix}%
\usepackage{xcolor}%
\usepackage{textcomp}%
\usepackage{manyfoot}%
\usepackage{algorithm,algorithmic}%
\usepackage{natbib}

\usepackage[nolist,nohyperlinks]{acronym}
\acrodef{AI}[AI]{Artificial Intelligence}
\acrodef{DTL}[DTL]{Deep Transfer Learning}
\acrodef{SNN}[SNN]{Spiking Neural Network}
\acrodef{ANN}[ANN]{Artificial Neural Network}
\acrodef{STDP}[STDP]{Spike-Timing-Dependent Plasticity}
\acrodef{MAC}[MAC]{Multiplication-Accumulation}
\acrodef{BPTT}[BPTT]{Backpropagation Through Time}
\acrodef{CNN}[CNN]{Convolutional Neural Network}
\acrodef{IF}[IF]{Integrate-and-Fire}
\acrodef{QCFS}[QCFS]{Quantization-Clip-Floor-Shift}
\acrodef{FA}[FA]{Feedback Alignment}
\acrodef{DFA}[DFA]{Direct Feedback Alignment}
\acrodef{LIF}[LIF]{Leaky Integrate-and-Fire}

\def\LET#1#2{\STATE #1 $\leftarrow$ #2}

\newcommand{\subfig}[3]{    \begin{subfigure}[b]{0.15\textwidth}
		\centering
		\includegraphics[width=\textwidth,keepaspectratio]{#1}
		\caption{#2}
		\label{#3}
	\end{subfigure}
}

\title{Energy-Efficient Eimeria Parasite Detection Using a Two-Stage Spiking Neural Network Architecture}

\author{
 Ángel Miguel García-Vico \\
  Andalusian Research Institute in Data Science and Computational Intelligence (DaSCI)\\
  University of Jaén\\
  Jaén, 23071, Spain\\
  \texttt{agvico@ujaen.es} \\
   \And
  Huseyin Seker\\
  Department of Information Systems, College of Computing and Informatics\\
  University of Sharjah\\
  Sharjah \\
  \texttt{hseker@sharjah.ac.ae} \\
  \And
 Muhammad Afzal \\
  Department of Computer Science, School of Architecture, Built Environment, Computing, and Engineering\\
  Birmingham City University\\
  Birmingham, United Kingdom \\
  \texttt{Muhammad.Afzal@bcu.ac.uk} \\
}

\begin{document}
\maketitle
\begin{abstract}
Coccidiosis, a disease caused by the Eimeria parasite, represents a major threat to the poultry and rabbit industries, demanding rapid and accurate diagnostic tools. While deep learning models offer high precision, their significant energy consumption limits their deployment in resource-constrained environments. This paper introduces a novel two-stage Spiking Neural Network (SNN) architecture, where a pre-trained Convolutional Neural Network is first converted into a spiking feature extractor and then coupled with a lightweight, unsupervised SNN classifier trained with Spike-Timing-Dependent Plasticity (STDP). The proposed model sets a new state-of-the-art, achieving 98.32\% accuracy in Eimeria classification. Remarkably, this performance is accomplished with a significant reduction in energy consumption, showing an improvement of more than 223 times compared to its traditional ANN counterpart. This work demonstrates a powerful synergy between high accuracy and extreme energy efficiency, paving the way for autonomous, low-power diagnostic systems on neuromorphic hardware.
\end{abstract}


\section{Introduction} \label{sec:intro}

Coccidiosis, a parasitic disease caused by protozoa of the genus Eimeria, represents one of the major threats to the poultry and rabbit industry worldwide. This disease mainly affects chickens, rabbits and other animals,  causing intestinal lesions, haemorrhagic diarrhoea, lethargy and, in acute cases, high mortality rates \cite{fatoba2018diagnosis, abdisa2019poultry, mesa2021chicken}. The parasite's complex life cycle and rapid spread through contaminated water and feed consumption seriously  jeopardise the health and productivity of these animals, resulting in significant economic losses for producers \cite{mesa2021chicken}. The need to develop faster and more practical diagnostic methods is therefore a fundamental pillar  for establishing effective control and prevention strategies that minimise the impact of the disease \cite{gao2024advancements}.

Traditional diagnosis of Eimeria species has historically been based on microscopic examination of oocysts, a process that presents significant challenges \cite{quinn2016deep,rajapaksha2019review,abdalla2018developing,kumar2023advances}. Accurate species identification is a complex task due to  the high morphological similarity between them. This method is not only laborious and time-consuming but also requires a high level of expertise from specialised technical staff, a resource that is not always  available. 
Nevertheless, in recent years, \ac{AI} has emerged as an innovative solution for the classification of microorganisms from microscopic images. These technologies facilitate the rapid recognition of pathogens, reducing dependence on qualified personnel and minimising the potential spread of infections. In the context of coccidiosis, deep learning models, especially those based on \ac{DTL}, have shown great potential for classifying Eimeria species with a high degree of accuracy \cite{monge2019classification, he2023reliable}, leveraging the knowledge of pre-trained models on large volumes of data to adapt them to this specific task.

However, despite their success in classification, conventional deep learning approaches often have high energy and computational demands. This factor can significantly limit their practical application, especially in resource-constrained environments such as remote farms or laboratories with modest equipment, where access to advanced hardware or a stable power supply is not guaranteed. 

Therefore, current research focuses on both improving the accuracy of models, and optimising their energy efficiency. This approach, aligned with the concept of Green AI \cite{10.1145/3381831}, seeks to develop sustainable and accessible solutions. In this regard, novel architectures such as \acp{SNN} \cite{maass1997networks}, which mimic the functioning of the human brain to process information with significantly lower energy consumption, offer a promising alternative. The combination of \acp{ANN} and \ac{SNN}s in hybrid models seeks to calibrate the balance between performance and efficiency \cite{kugele2021hybrid, kosta2023live, vazquez2024combining}. Altough this approach seems promising, it does not use the full energy-efficiency characteristics of \ac{SNN}s as some parts of the network are still \ac{ANN}-based.


This work presents a novel contribution that directly fills this gap, proposing a hybrid architecture that applies a new transfer learning approach. First, a pre-trained deep \ac{SNN} model, obtained via \ac{ANN}-to-\ac{SNN} conversion, is used as a feature extractor, or backbone. This backbone processes the image and generates a feature vector based on the output neural activity, which is then encoded into a new spike train using a Poisson encoder. Finally, these spikes serve as input to an unsupervised \ac{SNN} classifier which, based on competitive learning and lateral inhibition, learns to group the extracted features and assign a class to each image without the need for labels during its training. In short, this work proposes:

\begin{enumerate}
	\item  A purely spiking end-to-end classification model that offers an excellent balance between accuracy and energy consumption.
	\item A new transfer learning approach for \ac{SNN}s that relies on an unsupervised classification layer, opening up new opportunities for the development of more efficient and autonomous \ac{AI} systems.
\end{enumerate}

This paper is organised as follows: Section \ref{sec:background_related} presents the main concepts and related works used throughout this work. An in-depth explanation of the main components of the proposed method are presented in Section \ref{sec:method}. The experimental study, together with the results and its discussion are described in Section \ref{sec:study}. Finally, the conclusions of this work are presented in Section \ref{sec:conclusions}.

%
%
\section{Related Works} \label{sec:background_related}

The accurate and rapid identification of Eimeria parasites is a critical challenge with significant economic and public health implications. Historically, this process relied on manual microscopic examination, a method fraught with difficulties due to the morphological similarity among species, making it heavily dependent on expert technicians \cite{quinn2016deep,rajapaksha2019review}. This has driven a shift towards automated, computational solutions \cite{kumar2023advances}. Early attempts involved traditional machine learning with manual feature extraction \cite{zhang2022comprehensive,men2008application,liu2001cmeias}. However, the advent of deep learning, particularly \acp{CNN}, revolutionised the field by enabling automatic learning of hierarchical features from raw image data \cite{zhang2021lcu, he2023reliable}.

A major hurdle in this specialised domain is the scarcity of large annotated datasets. \ac{DTL} has become the standard solution, reusing knowledge from models pre-trained on large-scale datasets like ImageNet to achieve high performance on smaller, specific datasets \cite{iman2023review,tan2018survey}. The effectiveness of \ac{DTL} in Eimeria classification is well-documented. For instance, a comparative study found that an Xception-based feature extractor achieved 96.4\% accuracy \cite{kucukkara2025classification}, while the hybrid ResTFG model reached a 96.9\% by combining \ac{CNN} and Transformer architectures to capture both local and global features \cite{he2023reliable}. Finally, the state-of-the-art is the work presented in \cite{acmali2024green} by means of a prunned version of the EfficientNet-B0 network, achieveing an accuracy of 95.1\% on chicken and 97.4\% on the rabbit dataset.

Despite their accuracy, these conventional \ac{ANN} models are computationally intensive, limiting their use in resource-constrained field settings. This has spurred interest in Green AI \cite{10.1145/3381831} and energy-efficient alternatives like \acp{SNN} \cite{maass1997networks}. Unlike \acp{ANN}, \acp{SNN} process information using discrete, asynchronous spikes, which drastically reduces energy consumption by replacing expensive \ac{MAC} operations with simpler additions \cite{kucik2021investigating}. However, training deep \acp{SNN} is challenging due to the non-differentiable nature of spike events. The research community has developed three main approaches to address this \cite{garcia2021preliminary}: direct training with surrogate gradients \cite{zenke2018superspike}, unsupervised learning with bio-inspired rules like \ac{STDP} \cite{Diehl2015-vb}, and the conversion of pre-trained \acp{ANN} to \acp{SNN} \cite{rueckauer2017conversion}.

\ac{ANN}-\ac{SNN} conversion is a popular and effective strategy that leverages mature \ac{ANN} training pipelines. The core idea is to map the continuous activation of an \ac{ANN} neuron to the firing rate of a spiking neuron. Early methods suffered from accuracy loss and high latency \cite{cao2015spiking,diehl2015fast}, but significant advances have been made. Techniques such as soft reset mechanisms \cite{rueckauer2017conversion,han2020rmp} and quantization-aware training using functions like \ac{QCFS} \cite{bu2023optimal} have been introduced to minimise conversion errors and reduce the required simulation timesteps. These optimisations have made conversion a viable path for creating high-performance, deep \acp{SNN}.

On the other hand, \ac{STDP} offers a local, unsupervised, and biologically plausible learning mechanism where synaptic weight is adjusted based on the precise timing of pre- and post-synaptic spikes \cite{song2000competitive}. Networks using \ac{STDP}, combined with other mechanisms such as lateral inhibition, can self-organise to perform complex recognition tasks without labels, making them highly suitable for on-chip learning in neuromorphic hardware \cite{Diehl2015-vb}. While hybrid methods combining \ac{STDP} with supervised signals exist \cite{lillicrap2016random,tavanaei2019bp,mazurek2025three}, the pure unsupervised approach remains attractive for its efficiency.

To date, the application of \acp{SNN} to Eimeria parasite detection remains largely unexplored. The only exception is a hybrid ANN-SNN model that used a ViT feature extractor with an SNN classifier trained via surrogate gradients, achieving up to 87.6\% accuracy, while reducing its energy comsumption by 60\% \cite{vazquez2024combining}. However, this approach is not fully spiking and it is not able to leverage the full energy efficiency of such models as the ViT represents a high percentage of the whole network. This work fills this gap by proposing a novel, purely spiking architecture that combines a deep SNN backbone, optimised via conversion, with an unsupervised STDP-based classifier, creating a new transfer learning paradigm for energy-efficient and autonomous diagnostic systems.

%
%

\section{Proposed method} \label{sec:method}

In this paper, a two-stage training methodology that combines supervised and unsupervised learning to create an efficient and accurate Eimeria classification system based on \ac{SNN}s is proposed. The first stage specializes a deep \ac{SNN} backbone, obtained via ANN-\ac{SNN} conversion, for feature extraction. In the second stage, this backbone is frozen and coupled to an unsupervised classifier based on the \ac{STDP} learning rule, delegating the final classification to a local and efficient plasticity mechanism.

\subsection{Network Architecture} \label{sec:arch}

The proposed architecture is a composite system that consists of a convolutional spiking backbone, a supervised temporal classifier for the first stage, and an unsupervised \ac{STDP} classifier for the second stage.

\subsubsection{Spiking convolutional backbone} \label{sec:stage1}

The core of the entire system is the backbone model, which is a deep convolutional architecture adapted to operate in the spiking domain. As a starting point, 25 different well-known high-performance convolutional architectures are used, such as ResNet-18 \cite{he2016deep}, VGG-11 \cite{simonyan2014very}, among others, which are pre-trained on ImageNet. The transfer learning strategy is essential for initialising the model with a generic and robust visual feature extractor. 

In order to carry out this process, an ANN-SNN conversion is performed to avoid significant performance loss by the backbone in the spiking domain. This conversion requires two critical structural modifications, which are key to enabling the network to work with spikes properly:

\begin{enumerate}
	
	\item Replacing MaxPooling with AveragePooling: MaxPooling layers are replaced with AveragePooling to avoid information loss in the firing rate of neurons, thus preserving a more faithful representation of the neural activity crucial for the spiking model.
	
	\item Replacement of activation functions by \ac{IF} Neurons:  All the activation functions within the backbone are replaced by spiking neurons. Specifically, these are replaced by an implementation of an \ac{IF} neuron whose behaviour is based on the \ac{QCFS} activation function \cite{bu2023optimal} to optimise the conversion from ANN to SNN. The \ac{QCFS} activation function maps the continuous activations of the ANN to a discrete number of firing levels, which significantly reduces the quantisation error inherent in the conversion. In addition, the function incorporates a shift term that minimises the expected conversion error when the number of simulation time steps does not match the number of quantisation levels ($L$) of the neuron.
	
\end{enumerate}

These modifications significantly alter the network's dynamics, so the pre-trained weights from ImageNet are no longer optimal. Consequently, a slight re-training of the backbone with our data is crucial to maximize its feature extraction capability in the new spiking domain.

\subsubsection{Classification header} \label{sec:classification}

Depending on the training stage, two different classification headers are used:

\begin{enumerate}
	\item Supervised classifier in the first stage.   For the first stage of fine-tuning, i.e. to readjust the weights of the backbone for the spiking domain, a classification header is added to the backbone. It consists of a sequence \texttt{Linear} $\rightarrow$ \texttt{IF} $\rightarrow$ \texttt{Linear}. This structure maintains the computational consistency of the spike domain up to the output layer and allows supervised training of the entire network using backpropagation by means of the \ac{QCFS} function.
	
	\item Unsupervised STDP-based classifier in the second stage. It consists of two layers: an input layer ($I$) and a processing layer ($R$). The STDP learning rule is specifically applied to the synapses of the dense, excitatory feed-forward connection from layer $I$ to layer $R$, enabling the neurons in layer $R$ to learn and become selective to recurring input patterns in an unsupervised manner. The processing layer $R$ contains adaptive \ac{LIF} neurons. Their adaptability comes from dynamically adjusting each neuron's firing threshold. When a neuron fires, its threshold increases by a given amount $\theta_+$ and then exponentially decays back towards the baseline. This makes it temporarily harder for a very active neuron to fire again, which encourages that all neurons participate in the learning process and prevents a few from dominating the network activity. Finally, a layer of lateral inhibition is established by inhibitory connections within layer $R$, where each neuron connects to all others but itself. This creates competition among the excitatory neurons, forcing them to specialize and respond to different features of the input data.
	
\end{enumerate}

\subsection{Neural encoding} \label{sec:coding}

In order to use the \ac{SNN}, it is necessary to convert the inputs, consisting of real numbers, into spike trains. In this work, two encoding strategies are employed. Firstly, encoding is carried out in the backbone by means of a direct input encoding strategy \cite{rathi2021diet}. Here, the normalised input pixel values are directly injected into the network as input current. In this way, the encoding to spike trains is carried out by the first \ac{IF} layer of the backbone. This strategy has been widely used in the literature due to its good classification results \cite{zhou2024direct, yao2025scaling}. 

On the other hand, as detailed in Section \ref{sec:training}, the output of the backbone is the average firing rate. In order to subsequently use the \ac{STDP} classifier, a second conversion to spike trains is carried out using the Poisson encoder \cite{van2001rate}, which has proven to be a robust encoding method for rate coding \cite{sharmin2020inherent}.

Finally, given that the Stage 2 classifier is trained in an unsupervised manner, a method is required to interpret its activity. First, a label assignment process is carried out, where each neuron is assigned the label of the class that most frequently causes it to fire during a calibration phase. To refine the final prediction, the proportion of times a neuron fires for its assigned class compared to its total firing activity across all classes is computed. This allows to calculate how specialised is the neuron for a specific stimuli for that class. For the final prediction, the activity of each neuron is then weighted by this value. This proportion weighting method offers a more reliable prediction than a simple spike count, as it prioritizes the activity of neurons highly specialized for a specific class.

\subsection{Training procedure} \label{sec:training}

The training of the proposed model is carried out in two main stages, which are formalized in Algorithms \ref{alg:training} and \ref{alg:training2}. The first stage consists of the supervised adaptation of the backbone, while the second stage focuses on the unsupervised training of the STDP-based classifier.

\begin{figure*}[t!]
	\centering
	\begin{subfigure}[t]{0.4\textwidth}
		\centering
		\includegraphics[width=\linewidth]{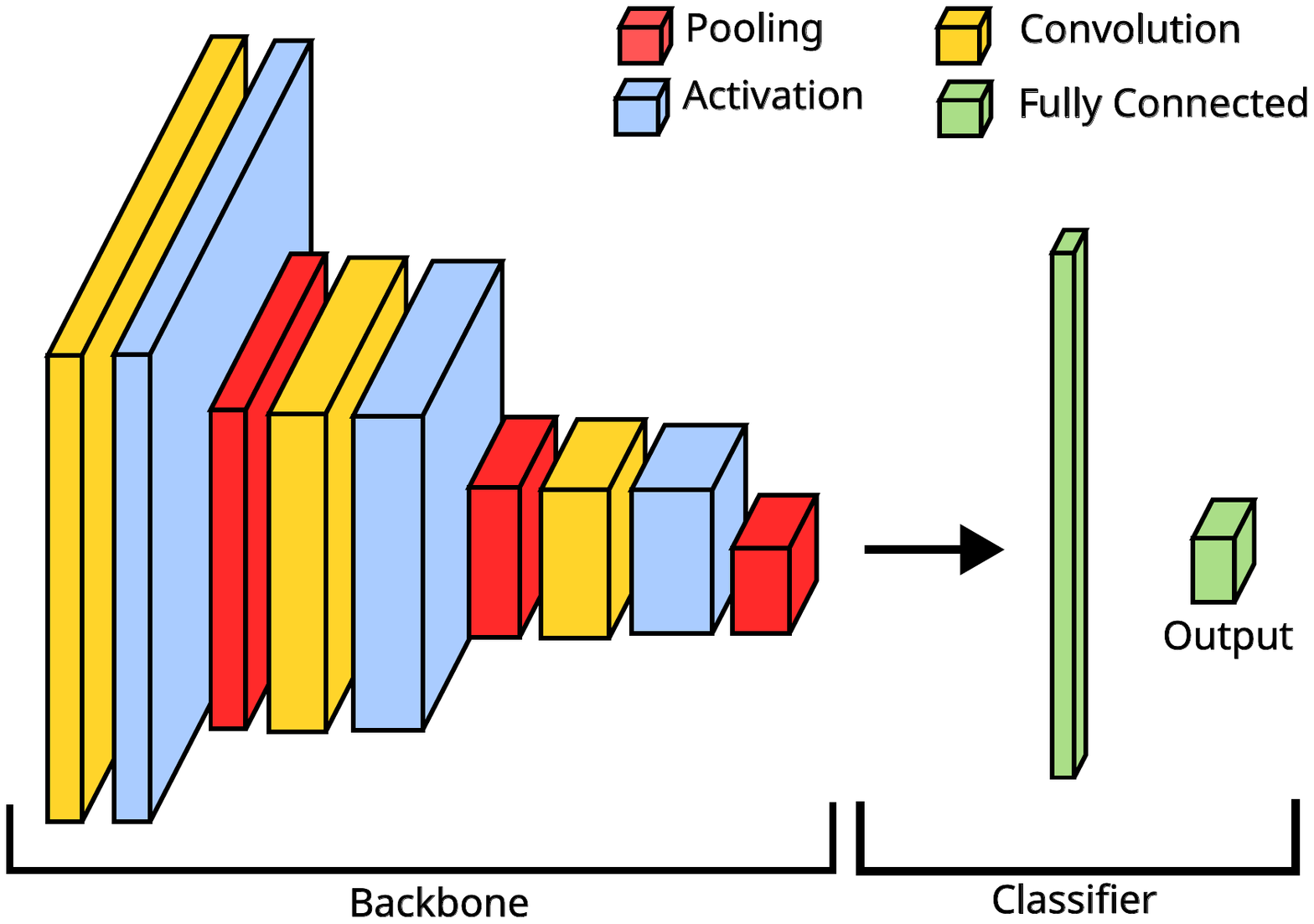}
		\caption{Initial Setup: Classic finetuning}
		\label{fig:initial_state}
	\end{subfigure}%
	~ 
	\begin{subfigure}[t]{0.4\textwidth}
		\centering
		\includegraphics[width=\linewidth]{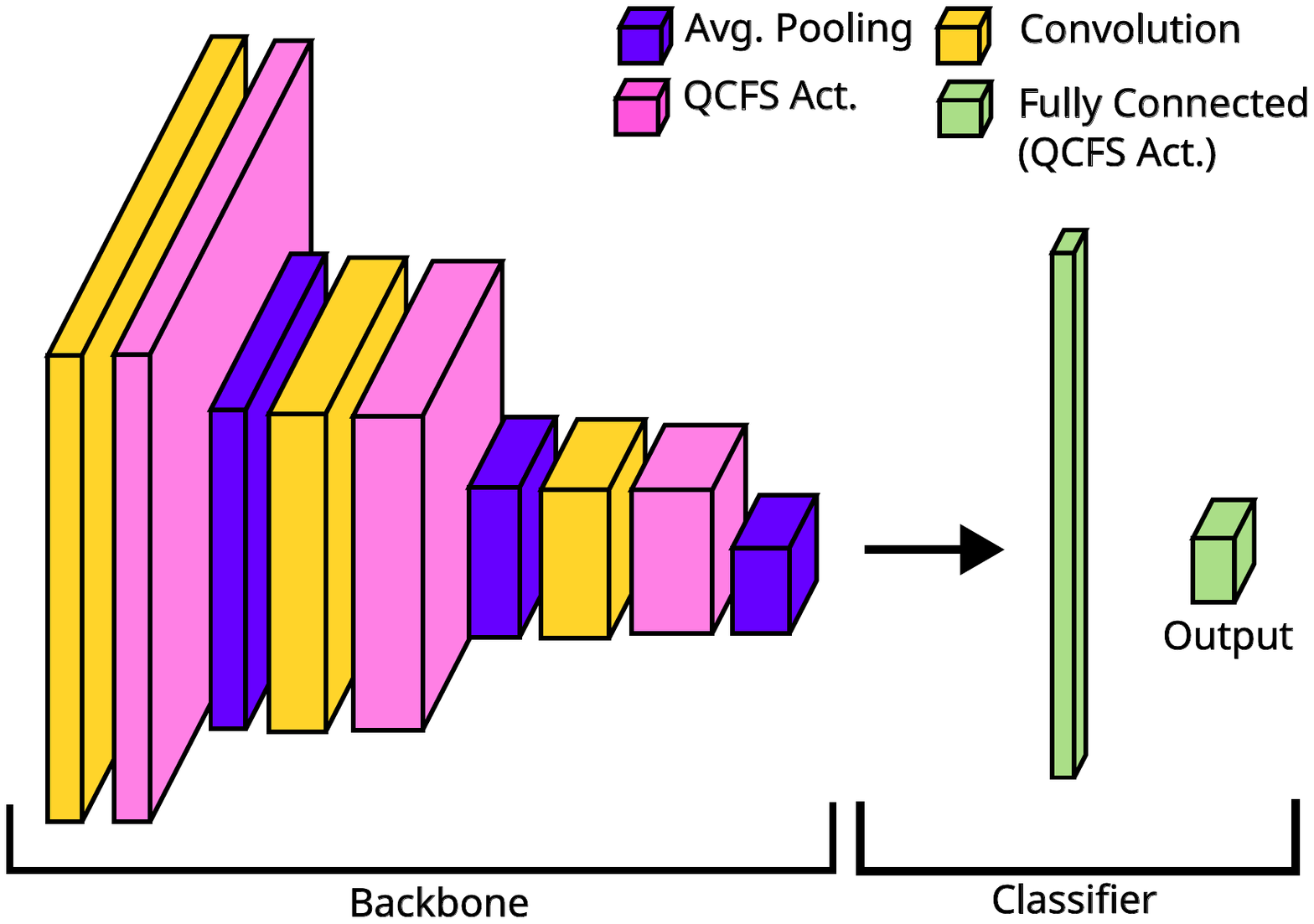}
		\caption{Stage 1.}
		\label{fig:stage2}
	\end{subfigure}
	
	\begin{subfigure}[b]{0.4\textwidth}
		\centering
		\includegraphics[width=\linewidth]{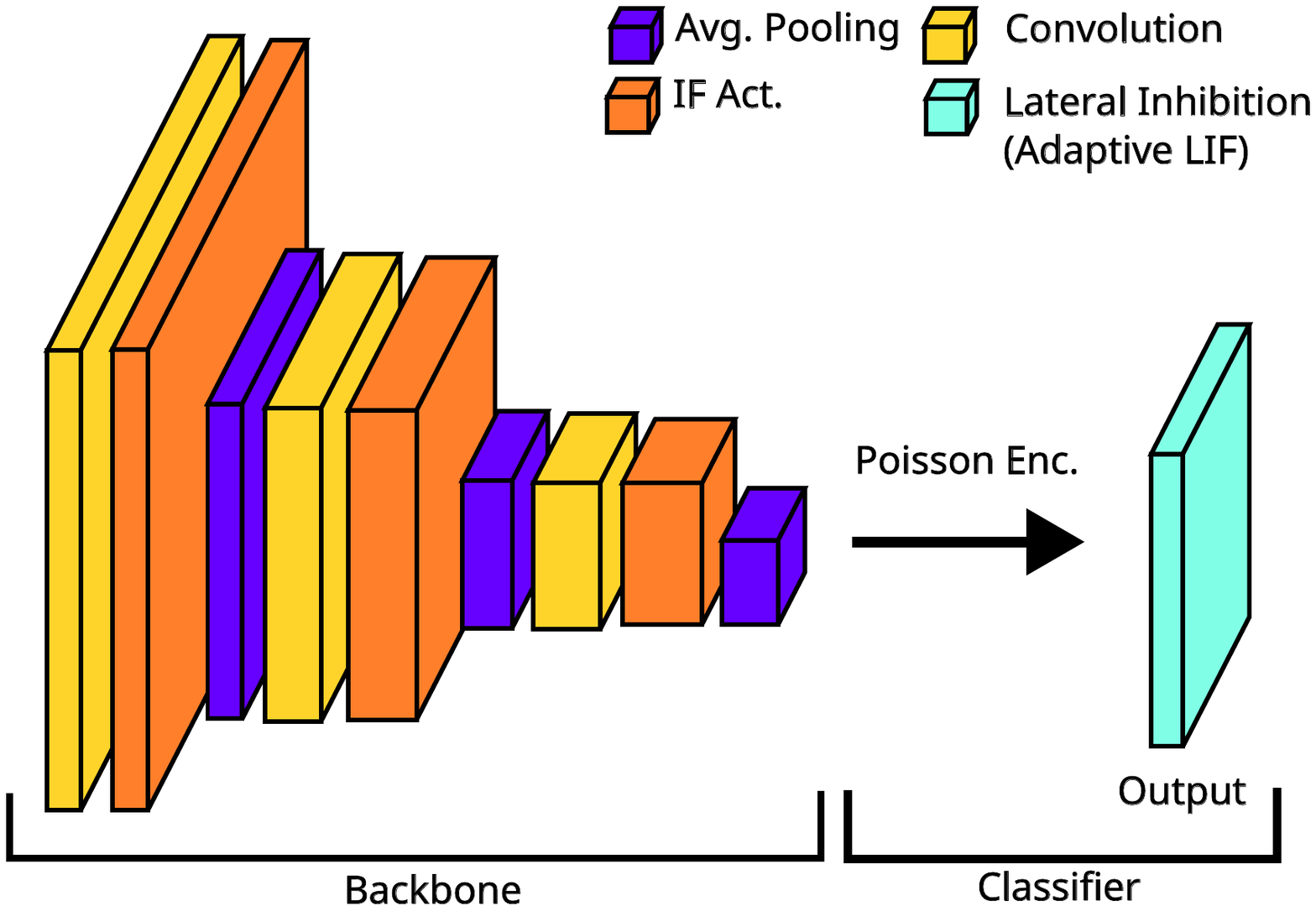}
		\caption{Stage 2. Proposed final model.}
		\label{fig:stage3}
	\end{subfigure}
	\caption{Proposed two-stages training procedure. The process start with a classical fine-tuning approach, as shown in (a). In the Stage 1, shown in (b), activations functions are replaced by the QCFS activation and max-pooling is replaced by average pooling. Full-finetuning is carried out. The final Stage 2 (c) is carried out where the classification header is discarde and replaced by the unsupervised STDP classifier. The backbone's weights are frozen and only the classifier is trained by means of STDP.}
	\label{fig:stages}
\end{figure*}

The initial phase involves adapting the pre-trained backbone for the spiking domain, as illustrated in the transition from a classic fine-tuning setup (Fig. \ref{fig:initial_state}) to the SNN-ready architecture (Fig. \ref{fig:stage2}). To ensure compatibility with the future SNN conversion, two critical modifications are made prior to training: all standard activation functions are replaced with the \ac{QCFS} function, and all MaxPooling layers are substituted with Average Pooling layers (Algorithm \ref{alg:training}, lines 1-7). Since these alterations can degrade initial performance, a complete retraining is necessary. To this end, the entire model is fine-tuned using the Cross-Entropy loss function with a cosine annealing scheduler (lines 9-20). A key aspect of this process is the concurrent optimization of both the network's synaptic weights ($W$) and the firing thresholds ($\lambda$) of the QCFS function (lines 15-16). This allows the network to find an optimal balance between information transmission and energy efficiency. Upon completion, the trained ANN is converted into an SNN: the synaptic weights are inherited directly, the firing threshold of the \ac{IF} neurons is set from the optimized $\lambda$ values, and the initial membrane potential is set to half the threshold, a heuristic that minimizes conversion error \cite{bu2023optimal} (lines 21-23).

\begin{algorithm}[!hbtp]
	\caption{Training procedure for Stage 1.}
	\label{alg:training}
	\scriptsize
	
	\hspace*{\algorithmicindent} \textbf{Input:} \\
	\hspace*{\algorithmicindent}\hspace*{\algorithmicindent} $L$: Quantization step for the QCFS funcition. \\ 
	\hspace*{\algorithmicindent}\hspace*{\algorithmicindent} $\lambda$: Starting firing threshold for the IF neuron. \\
	\hspace*{\algorithmicindent}\hspace*{\algorithmicindent} $\Theta_{ANN}$: The backbone model. \\
	\hspace*{\algorithmicindent}\hspace*{\algorithmicindent} $D$: Dataset. \\
	\hspace*{\algorithmicindent} \textbf{Output:} \\
	\hspace*{\algorithmicindent}\hspace*{\algorithmicindent} $\Theta_{SNN}$: A trained SNN backbone for Stage 2.
	
	\begin{algorithmic}[1]
		
		\FORALL{layers $l_i \in \Theta_{ANN}$}
		\IF{$l_i$ is an activation function}
		\LET{$l_i$}{QCFS(L, $\lambda_i$)}
		\ENDIF
		\IF{$l_i$ is MaxPooling layer}
		\LET{$l_i$}{AvgPooling}
		\ENDIF
		\ENDFOR
		
		\FOR{$e = 1$ to epochs}
		\FOR{length of $D$}
		\LET{($x$, $y$)}{Sample minibatch from $D$}
		\LET{$\hat{y}$}{$\Theta_{ANN}\left(x\right)$}
		\LET{Loss}{CrossEntropy($\hat{y}, y)$}
		\FORALL{layer $l \in \Theta_{ANN}$}
		\LET{$W^l$}{$W^l - \epsilon \frac{\partial \text{Loss}}{\partial W^l}$}
		\LET{$\lambda^l$}{$\lambda^l - \epsilon \frac{\partial \text{Loss}}{\partial \lambda^l}$}
		\ENDFOR
		\LET{$\epsilon$}{Cosine Annealing LR Scheduler}
		\ENDFOR
		\ENDFOR
		
		\LET{$\Theta_{SNN}.W$}{$\Theta_{ANN}.W$}
		\LET{$\Theta_{SNN}.\theta$}{$\Theta_{ANN}.\lambda$}
		\LET{$\Theta_{SNN}.v(0)$}{$\Theta_{SNN}.\theta/2$}
		
		\RETURN{$\Theta_{SNN}$}
	\end{algorithmic}
\end{algorithm}

In the second stage, the supervised classification layers are discarded and replaced with the unsupervised SNN classifier, as shown in Fig. \ref{fig:stage3}. The training process for this stage is detailed in Algorithm \ref{alg:training2}. The backbone network's weights are frozen to act as a static feature extractor, and the synaptic weights of the new classifier are randomly initialized (lines 1-2). For each input sample, the backbone generates a firing rate vector over a duration $T_b$ (line 5), which is then encoded into a spike train using a Poisson encoder for a duration $T_c$ (line 6). This spike train serves as the input for the unsupervised classifier. During the simulation over $T_c$ timesteps, the network adapts through several mechanisms: lateral inhibition fosters competition among neurons (line 12), allowing them to specialize; the STDP rule updates synaptic weights based on spike timing (line 14); and an adaptive threshold mechanism ($\theta_+$) encourages all neurons to participate in the learning process (line 15).

\begin{algorithm}[!hbtp]
	\caption{Stage 2: Unsupervised STDP-based training of the classifier.}
	\label{alg:training2}
	\scriptsize
	
	\hspace*{\algorithmicindent} \textbf{Input:} \\
	\hspace*{\algorithmicindent}\hspace*{\algorithmicindent} $exc$: Excitatory population firing threshold. \\ 
	\hspace*{\algorithmicindent}\hspace*{\algorithmicindent} $inh$: Inhibitory population firing threshold. \\ 
	\hspace*{\algorithmicindent}\hspace*{\algorithmicindent} $T_b$: Presentation time for the backbone feature extractor. \\ 
	\hspace*{\algorithmicindent}\hspace*{\algorithmicindent} $T_c$: Presentation time for the SNN classifier. \\ 
	\hspace*{\algorithmicindent}\hspace*{\algorithmicindent} $\theta_+$: Adaptive threshold increase value (homeostasis). \\
	\hspace*{\algorithmicindent}\hspace*{\algorithmicindent} $\Theta_{SNN}$: The pre-trained backbone SNN model. \\
	\hspace*{\algorithmicindent}\hspace*{\algorithmicindent} $D$: Training dataset. \\
	\hspace*{\algorithmicindent} \textbf{Output:} \\
	\hspace*{\algorithmicindent}\hspace*{\algorithmicindent} The trained model.
	
	\begin{algorithmic}[1]
		
		\STATE Freeze $\Theta_{SNN}$ parameters.
		\STATE Initialize synaptic weights $W_{I \rightarrow R}$ randomly.
		\FOR{$e = 1$ to epochs}
		\FOR{each sample ($x$, $y$) in $D$}
		\LET{$r_{f}$}{$\Theta_{SNN}(x, T_b)$} \COMMENT{Get firing rates from backbone}
		\LET{$S_{in}$}{$\text{PoissonEncoder}(r_f, T_c)$} \COMMENT{Encode features into input spike train} 
		
		\FOR{$t = 1$ to $T_c$}
		\LET{$v_{exc}$}{$v_{exc} + S_{in}(t)\cdot W_{I \rightarrow R}$} \COMMENT{Update membrane potential neurons}
		\LET{$S_{exc}(t)$}{$v_{exc} \geq \theta_{exc}$} \COMMENT{Check for excitatory spikes}
		\LET{$v_{inh}$}{$v_{inh} + S_{exc}(t) \cdot exc$} \COMMENT{Update inhibitory membrane potential}
		\LET{$S_{inh}(t)$}{$v_{inh} \geq \theta_{inh}$} \COMMENT{Check for inhibitory spikes}
		\LET{$v_{exc}$}{$v_{exc} - S_{inh} \cdot inh \cdot W_{R \rightarrow R}$} \COMMENT{Lateral inhibition}
		\STATE Reset potential $v_{exc}$ for neurons that fired.
		\STATE Update weights $W_{I \rightarrow R}$ based on the timing of $S_{in}(t)$ and $S_{exc}(t)$ (STDP rule).
		\LET{$\theta_{exc}$}{$\theta_{exc} + \theta_+$} \COMMENT{Adapt firing thresholds for neurons that fired}
		\ENDFOR
		\ENDFOR
		\ENDFOR
		
	\end{algorithmic}
\end{algorithm}

Finally, after the unsupervised training is complete, a label assignment procedure is executed. Each neuron is assigned the class label that elicits its maximum firing rate during training. For inference,
the final classification is determined by weighting the activity of each neuron by its specialization to its assigned class, providing a more robust prediction than a simple spike count.

\section{Experimental study} \label{sec:study}

In this section, the experimentation is detailed. First, the experimental framework where datasets and methods employed and the experimental pipeline are described in Section \ref{sec:framework}. Next, the method for measuring energy consumption is outlined in Section \ref{sec:energy}. Finally, the results obtained together with a detailed analysis is depicted at Section \ref{sec:results}.

\subsection{Experimental framework} \label{sec:framework}

The experimental study is conducted on a Linux system running Ubuntu 24.04 LTS, with 64 GB of RAM, AMD Ryzen 9 9950X 16-Core Processor at 5.7 Ghz, and a NVIDIA RTX 5070 GPU with 12GB of VRAM.

The remaining information about this section is divided into different subsections: Datasets where description and properties about data are shown, the different methods used as backbone and as a classifier together with their properties, and finally, the experimental pipeline for this study is shown.

\subsubsection{Datasets}

The data for this study were sourced from two distinct datasets provided by [32]. Both datasets are composed of images featuring oocysts of various microparasite species from the genus Eimeria, which infect either chickens or rabbits. The composition of each dataset, including the specific species and the quantity of images per species, is detailed in Table \ref{tab:dataset_distribution}.

\begin{table}[hbtp!]
	\centering
	\scriptsize
	\caption{Distribution of Images by Species in Chicken and Rabbit Datasets.}
	\label{tab:dataset_distribution}
	\setlength{\tabcolsep}{4pt} 
	\begin{tabular}{lr|lr}
		\hline
		\multicolumn{2}{c}{\textbf{Chicken Dataset}} & \multicolumn{2}{c}{\textbf{Rabbit Dataset}} \\
		\textbf{Species} & \textbf{Images} & \textbf{Species} & \textbf{Images} \\
		\hline \hline
		E. Acervulina & 742 & E. Coecicola & 170 \\
		E. Brunetti & 442 & E. Exigua & 283 \\
		E. Maxima & 360 & E. Flavescens & 292 \\
		E. Mitis & 820 & E. Intestinalis & 521 \\
		E. Necatrix & 502 & E. Irresidua & 213 \\
		E. Praecox & 898 & E. Magna & 572 \\
		E. Tenella & 696 & E. Media & 210 \\
		& & E. Perforans & 315 \\
		& & E. Piriformis & 136 \\
		& & E. Stiedai & 229 \\
		& & E. Vejdovski & 296 \\
		\hline
		\textbf{Total} & \textbf{4460} & \textbf{Total} & \textbf{3237} \\
		\hline
	\end{tabular}
\end{table}

\begin{figure}[htbp!]
	\centering
	\subfig{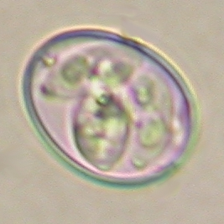}{Acervulina}{fig:acervulina}
	\hfill%
	\subfig{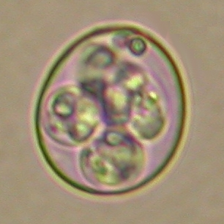}{Brunetti}{fig:brunetti}
	\hfill%
	\subfig{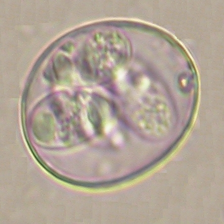}{Maxima}{fig:maxima}
	\hfill%
	\subfig{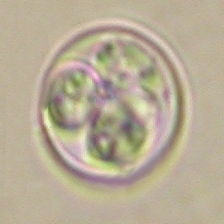}{Mitis}{fig:mitis}
	\hfill%
	
	\subfig{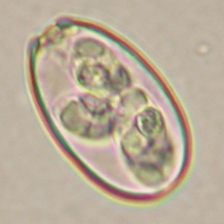}{Coecicola}{fig:coecicola}
	\hfill%
	\subfig{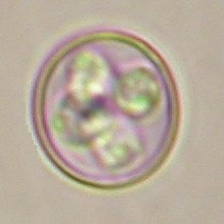}{Exigua}{fig:exigua}
	\hfill%
	\subfig{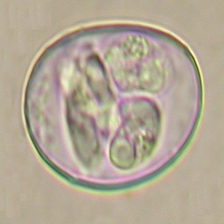}{Flavescens}{fig:flavescens}
	\hfill%
	\subfig{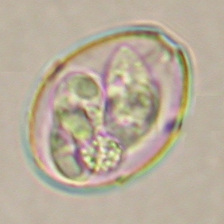}{Intestinalis}{fig:intestinalis}
	\hfill%
	
	\caption{Images of different species of Eimeria parasites. \subref{fig:acervulina}), \subref{fig:brunetti}), \subref{fig:maxima}) and \subref{fig:mitis}) correspond to classes for the chicken datasets, whereas \subref{fig:coecicola}), \subref{fig:exigua}), \subref{fig:flavescens}) and \subref{fig:intestinalis}) corresponds to classes for the rabbit dataset.}
	\label{fig:dataset}
\end{figure}

The images were produced under conditions representative of a typical use-case, utilizing a Nikon Eclipse E800 microscope and a Nikon Coolpix 4500 camera at 40x magnification. The resulting files were stored in a 24-bit JPEG format. To ensure compatibility with lower-resolution systems, the images were resized from their original 2272 x 1704 resolution to 224 x 224 pixels. A significant challenge, illustrated in Fig. \ref{fig:dataset}, is the morphological similarity across species.

\subsubsection{Methods}

In this study, a diverse set of pre-trained computer vision models were systematically evaluated to determine the best-performing architecture for classifying the Eimeria species. The performance of these models was assessed both \ac{ANN} format and after their conversion into \ac{SNN}. For the classifier layer, 500 neurons have been used for both stages.
This diversity of architectures allows for a comprehensive evaluation of classification performance and computational efficiency. A summary of the evaluated models and their key characteristics is presented in Table \ref{tab:model_architectures}.


\begin{table}[!hbtp]
	\centering
	\scriptsize
	\caption{Characteristics of the pre-trained architectures examined within the scope of the study. The number of parameters is shown in millions (M).}
	\label{tab:model_architectures}
	\begin{tabular}{lr}
		\hline
		\textbf{Network} & \textbf{Params (M)} \\ \hline \hline
		MNASNet-0.5 & 2.2 \\
		MobileNetV3-Small & 2.5 \\
		MNASNet-0.75 & 3.2 \\
		MobileNetV2 & 3.5 \\
		MNASNet-1.0 & 4.4 \\
		EfficientNet-B0 & 5.3 \\
		MobileNetV3-Large & 5.5 \\
		MNASNet-1.3 & 5.9 \\
		EfficientNet-B1 & 7.8 \\
		DenseNet-121 & 8.0 \\
		EfficientNet-B2 & 9.2 \\
		ResNet-18 & 11.7 \\
		EfficientNet-B3 & 12.0 \\
		DenseNet-169 & 14.1 \\
		EfficientNet-B4 & 19.0 \\
		EfficientNetV2-S & 21.0 \\
		ResNet-34 & 21.8 \\
		ResNet-50 & 25.6 \\
		EfficientNet-B5 & 30.0 \\
		ResNet-101 & 44.5 \\
		EfficientNetV2-M & 54.0 \\
		ResNet-152 & 60.2 \\
		ViT-B/16 & 86.0 \\
		VGG-11 & 133.0 \\
		VGG-16 & 138.0 \\ \hline
	\end{tabular}
\end{table}

\subsubsection{Experimental pipeline}

The experimental procedure was structured into three main phases: supervised backbone adaptation, unsupervised classifier training with hyperparameter optimisation, and final evaluation. Prior to training, the dataset was partitioned into training, validation, and test subsets using a stratified hold-out methodology. Initially, 20\% of the data was allocated to the test set. The remaining 80\% was further subdivided, with 10\% of this portion designated for validation and the other 90\% reserved for training. Stratification was performed based on parasite species during both splits to ensure the class distribution was maintained across all three subsets.

Also, to enhance the model's generalization capabilities and mitigate overfitting, a data augmentation protocol was applied exclusively to the training images. This process involved a series of stochastic transformations, including random horizontal flips, random rotations within a $\pm$ 20-degree range, and random adjustments to image brightness, contrast, and saturation with a factor of 0.2. The validation and test sets were not augmented, ensuring an unbiased evaluation of the model's performance.

After that, the supervised backbone adaptation stage begins. As described in Algorithm \ref{alg:training} the primary goal was to fine-tune the pre-trained architectures, listed in Table \ref{tab:model_architectures} to specialise them for the Eimeria dataset while accommodating the SNN-compatible architectural changes. This was carried out by training each modified backbone with its supervised classifier head, as shown in Section \ref{sec:classification}, in ANN-mode, i.e. $T_b=0$. The training process utilised the Adam optimiser, a Cosine Annealing learning rate scheduler, and the cross-entropy loss function, whose parameters are depicted in Table \ref{tab:snn_params}. The best version of each backbone was saved. This best version is based on its performance on the validation set. It is important to note that a classic fine-tuning approach without any modification was also performed as a baseline.

\begin{table}[hbtp!]
	\centering
	\scriptsize
	\caption{Hyperparameters for the supervised backbone adaptation stage.}
	\label{tab:snn_params}
	\begin{tabular}{lr}
		\hline
		\textbf{Hyperparameter} & \textbf{Value} \\
		\hline
		\hline
		Epochs & 100 \\
		Batch Size & 32 \\
		Optimizer & Adam \\
		Learning Rate & 1e-3 \\
		Cosine Annealing T\_max & 100 \\
		L Quantization & 8 \\
		IF Threshold & 2.0 \\
		\hline
	\end{tabular}
\end{table}

Next, the main phase of the study was the unsupervised training of the hybrid classifier described in Section \ref{sec:classification} and in Algorithm \ref{alg:training2}. In this stage, the best-performing backbone from the previous phase was loaded, and its weights were frozen to serve as a static feature extractor. For each image in the training set, the backbone generated a feature vector, which represents the firing rate of the output neurons. The rate is materialised by creating a sequence of spikes using a Poisson encoder following the firing rate of these neurons. These spike trains served as the input for the unsupervised SNN classifier which learns through STDP and lateral inhibition. After this unsupervised learning phase, labels were assigned to the classifier's neurons based on their firing responses to the labelled training data.

It is important to remark, that in order to determine the optimal configuration for this second stage, an extensive hyperparameter search based on bayesian search was conducted using the Optuna framework \cite{optuna_2019}. This process aimed to maximise classification accuracy by tuning key parameters, such as the increment in the threshold of the LIF neurons ($\theta_+$), and excitatory and inhibitory connection strengths (exc, inh). Here, minibatch training is carried out in order to speed-up training using a reduction on the minibatch dimension based on the maximum neurons activation \cite{saunders2019minibatch}. Also, an early stopping criterion is employed.

Once the optimal hyperparameters were identified, the model was trained a final time with these parameters on the combined training and validation sets, and its performance was evaluated on the held-out test set to produce the final results. As in the previous analysis, the best performing model on the validation set is stored to be used in the testing phase. The hyperparameters used in this part of the study are shown in Table \ref{tab:stdp_params}. Note that all the source code employed for this experimental study is avaible at GitHub\footnote{https://github.com/agvico/2025\_hybrid\_snn\_chicken\_parasites} for verification and reusability.

\begin{table}[h!]
	\centering
	\scriptsize
	\caption{Hyperparameters for the unsupervised STDP classifier training. Note that those values surrounded in square brackets are the minimum and maximum value for the hyperparameters search.}
	\label{tab:stdp_params}
	\begin{tabular}{ll}
		\hline
		\textbf{Hyperparameter} & \textbf{Value} \\
		\hline \hline
		inh & [150.0, 250.0]  \\
		exc & [20.0, 50.0]  \\
		$\theta_+$ & [0.001, 0.02] \\
		Batch Size & 32 \\
		Epochs & 10 \\
		Earyl stopping patience & 3 \\
		STDP learning rates & (1e-5, 1e-3) \\
		\hline
	\end{tabular}
\end{table}

\subsection{Energy consumption estimation} \label{sec:energy}

The energy consumption of the proposed method was theoretically estimated following the methodology outlined by Merolla et al. \cite{merolla2014million}. Accordingly, synaptic operations (SOPs) were considered the fundamental, energy-consuming operations for SNNs implemented on neuromorphic hardware. Within this framework, a SOP is contingent upon the generation of a spike. Therefore, the total spike count provides a reliable estimate of the synaptic operations performed by the SNN.

For the ANN, floating-point operations (FLOPs) were utilized as the basic computational unit. The total number of FLOPs was quantified using the \textit{thop}\footnote{https://github.com/ultralytics/thop} Python package. Based on the specifications of the ROLLS neuromorphic processor \cite{qiao2015reconfigurable}, we established energy consumption baselines of 77 fJ per SOP for the SNN and 12.5 pJ per FLOP for the ANN.

It is important to highlight that the energy costs associated with memory access were not considered in this analysis, as these are intrinsically dependent on the specific hardware architecture.

\subsection{Analysis of results} \label{sec:results}

The analysis of the results is structured in two sequential phases, mirroring the training methodology outlined in Section \ref{sec:training}. The first phase focuses on evaluating the supervised adaptation of the spiking backbones, while the second phase assesses the performance of the final hybrid model, which integrates the selected backbone with the unsupervised STDP classifier.

\subsubsection{Phase 1: Supervised Backbone Adaptation and Selection}

\begin{figure*}[t]
	\centering
	\begin{subfigure}[b]{0.4\textwidth}
		\centering 
		\includegraphics[width=\linewidth]{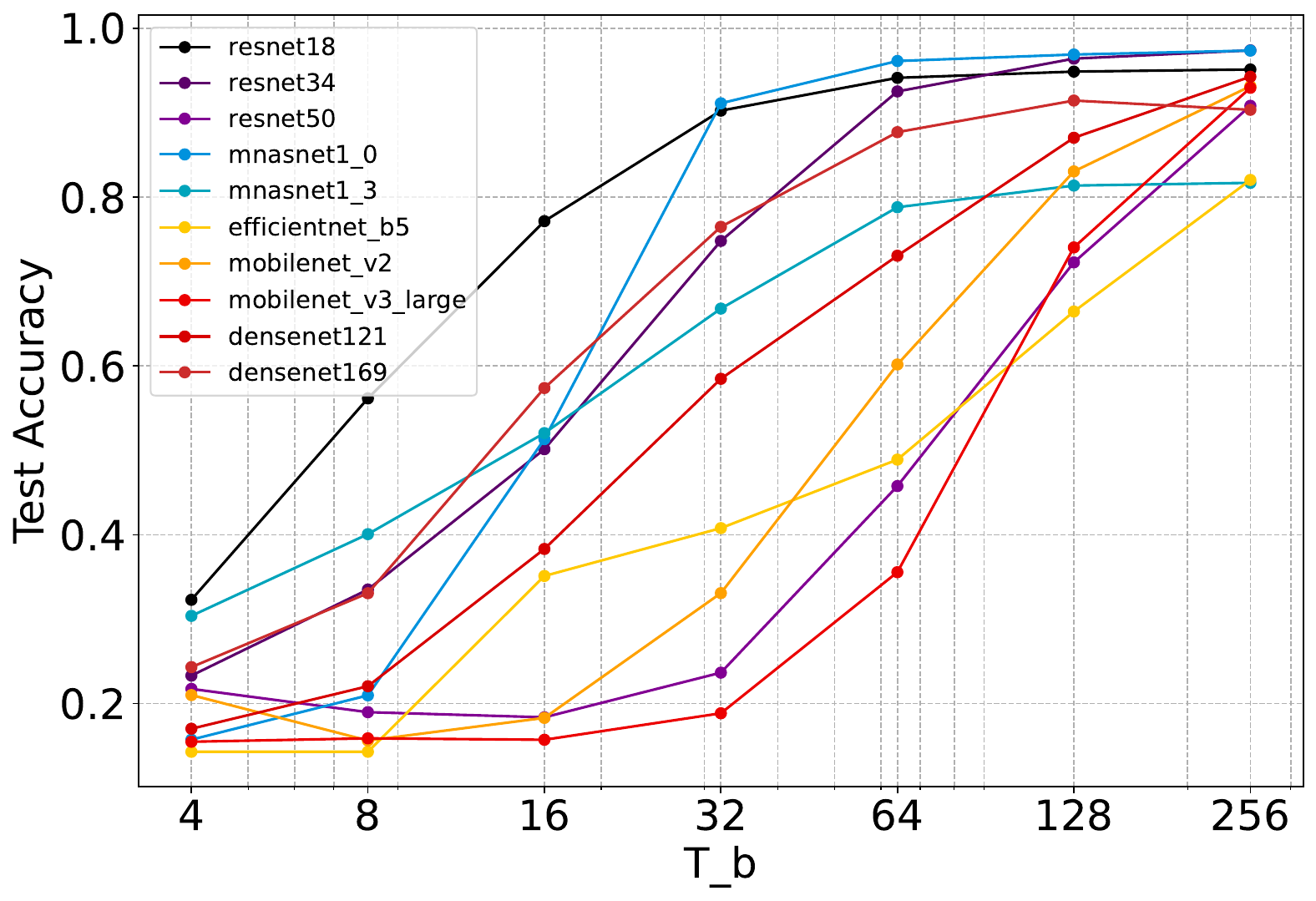}
		\caption{Chicken}.
	\end{subfigure}%
	~
	\begin{subfigure}[b]{0.4\textwidth}
		\centering 
		\includegraphics[width=\linewidth]{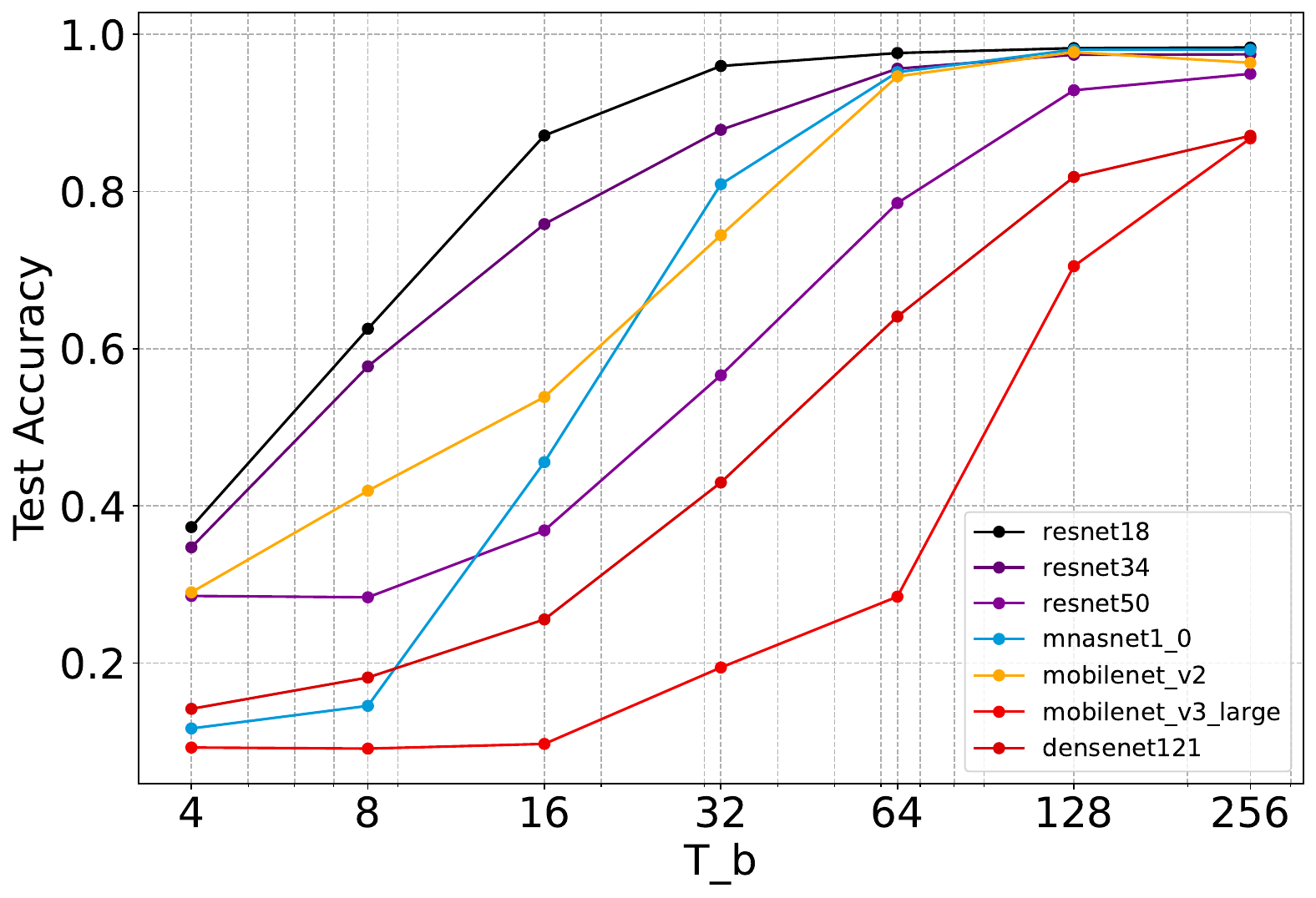}
		\caption{Rabbit}.
	\end{subfigure}
	\caption{Performance of the proposed backbones after the conversion process proposed in Section \ref{sec:stage1}. Note that only the models that achieved an accuracy higher than 80\% at $T_b = 256$ are shown. For the full table of results, please refer to Tables \ref{tab:phase1_chicken_results} and \ref{tab:phase1_rabbit_results}.}
	\label{fig:chickeStage1}
\end{figure*}

This analysis aimed to identify the highest-performing models. Fig. \ref{fig:chickeStage1} presents a graphical representation of the evolution of the proposed backbone methods along different presentation times ($T_b$) for both datasets. Nevertheless, it is important to remark that only the most promising models, i.e., those that achieved an accuracy higher than 80\% at $T_b = 256$, are shown in these figures for clarity. The complete result table for all methods analysed for both datasets is available at Table \ref{tab:phase1_chicken_results} for chickens and Table \ref{tab:phase1_rabbit_results} for rabbits.

The baseline ANN models ($T_b=0$), establish a strong performance ceiling, with most architectures achieving over 95\% accuracy on the chicken dataset and over 97\% on the rabbit dataset. This indicates a high level of feature learnability in both tasks. However, the conversion to the spiking domain reveals that at low simulation times ($T_b \le 16$), nearly all models experience a catastrophic drop in performance, often falling to the level of random guessing. This initial degradation is an expected consequence of the quantization and temporal coding errors inherent in the conversion process when the temporal resolution is insufficient. In fact, reducing this initial degration is nowadays an active research area.

Nevertheless, as simulation time increases, a clear divergence in architectural robustness becomes apparent. Certain model families demonstrate a remarkable ability to recover their baseline accuracy. The smaller variants of the ResNet family such as ResNet-18 or ResNet-34 emerge as exceptionally robust, consistently recovering to over 94\% accuracy on both datasets. Notably, on the rabbit dataset, they achieve this recovery faster, reaching over 95\% accuracy at just $T_b=64$. This suggests that their residual connections provide a stable path that translates well into the temporal dynamics of SNNs. Similarly, other less deep architectures such as MNASNet-1.0 and MobileNet-v2 also show strong resilience, particularly on the rabbit dataset, proving that efficiency-focused designs are not mutually exclusive with conversion robustness. 

In addition, it is remarkable the extreme brittleness of the EfficientNet family. Despite their high ANN accuracy, these models fail almost completely after conversion, showing negligible improvement with increased simulation time. This suggests that their core components, such as depth-wise separable convolutions and squeeze-and-excitation blocks, may be fundamentally incompatible with the chosen conversion method, losing critical information when translated to sparse spike trains. A similar failure is observed in the VGG and ViT models. 
In general, across both datasets, a performance elbow is evident around $T_b=128$. While some models on the rabbit dataset perform well at $T_b=64$, selecting $T_b=128$ ensures an optimal trade-off point for accuracy and latency. At this timestep, all successful models reach near-peak performance across both tasks without the significant additional computational cost of longer simulations, where returns are marginal.

Applying a performance threshold of 80\% accuracy at $T_b=128$, the most promising candidates for the final hybrid model were identified. The models that successfully passed this criterion on at least one dataset were selected, i.e., ResNet-18, ResNet-34, ResNet-50, MNASNet-1.0, MobileNet-v2, DenseNet-121 and DenseNet169. These architectures represent the most reliable candidates for the next phase, as they have proven their ability to generalize as effective spiking feature extractors for our problem.

\subsubsection{Phase 2: Performance of the Hybrid SNN-STDP Model}

In the second phase, the backbones selected from Phase 1 were integrated with the unsupervised STDP-based classifier. The primary goal here was to evaluate the performance of the complete end-to-end spiking model. After the hyperparameter optimization process described in Section \ref{sec:framework}, the final models were trained and evaluated.

\begin{figure*}[!hbtp]
	\centering
	\begin{subfigure}[b]{0.4\textwidth}
		\centering 
		\includegraphics[width=\linewidth]{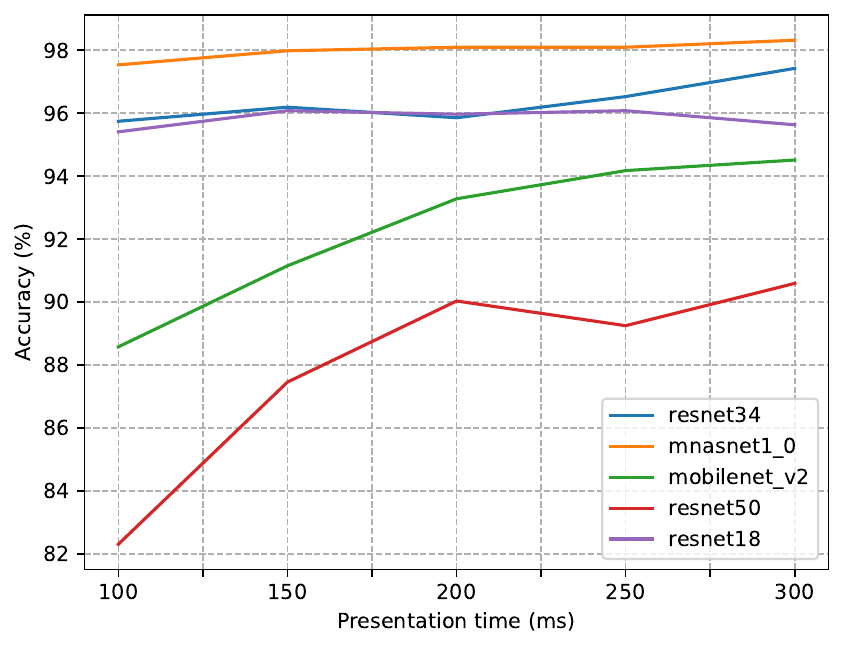}
		\caption{Chicken}.
	\end{subfigure}%
	~
	\begin{subfigure}[b]{0.4\textwidth}
		\centering 
		\includegraphics[width=\linewidth]{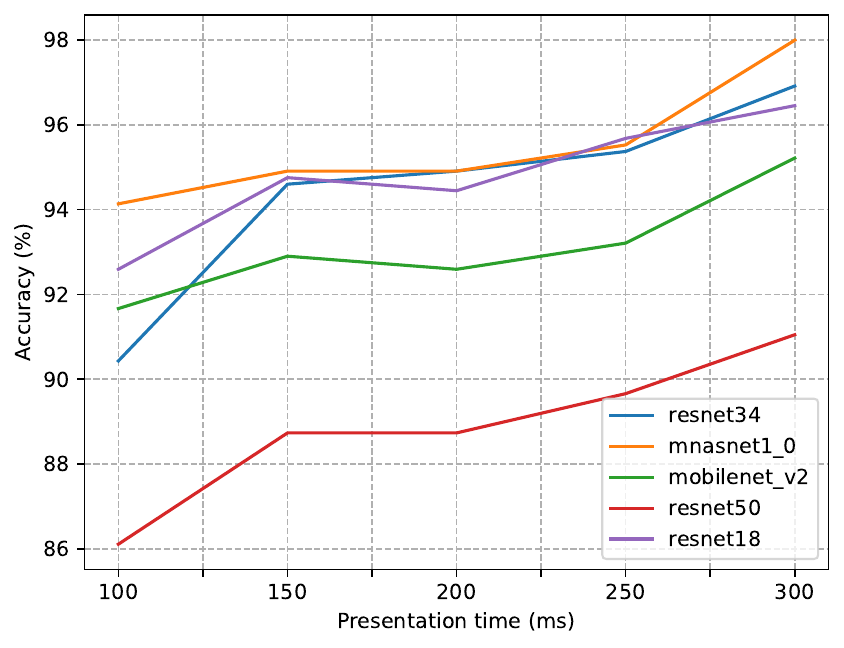}
		\caption{Rabbit}.
	\end{subfigure}
	\caption{Performance of the different backbones using the unsupervised STDP classification layer for several presentation times.}
	\label{fig:SNNs_timesteps_comparison}
\end{figure*}

To determine the optimal presentation time for the proposed hybrid architecture, the classifier's accuracy was evaluated across a range of timesteps, $T_c$, from 100 to 300 ms. The results for models achieving over 80\% accuracy on the target datasets are presented in Fig. \ref{fig:SNNs_timesteps_comparison}. The complete results can be found in the appendix  at Tables \ref{tab:SNN_accuracy_chicken} and \ref{tab:SNN_accuracy_rabbit}.

As can be observed, a clear trend emerges: classification accuracy generally improves as the presentation time $T_c$ increases. This is an expected behavior in rate-encoded SNNs, as longer integration times allow for more precise spike-rate approximation, leading to a more faithful representation of the feature embeddings. However, the models exhibit distinct performance profiles.

Models such as MNASNet-1.0, ResNet-34, and ResNet-18 demonstrate exceptional performance, achieving high accuracy even at short presentation times. For instance, on the Chicken dataset, MNASNet-1.0 reaches 97.54\% accuracy at just $T_c = 100$ ms, indicating a highly efficient processing of the data from the backbone. This suggests that a superior model achieved on Phase 1 directly translates to an SNN that requires fewer timesteps to converge to a robust solution, making these models ideal for low-latency applications.

Conversely, other models showcase a more pronounced trade-off between latency and accuracy. ResNet-50, for example, begins with a modest accuracy of 82.31\% at 100 ms but improves significantly to 90.59\% at 300 ms on the Chicken dataset. This behavior highlights a key insight: the proposed unsupervised SNN classifier demonstrates increasing robustness to lower-quality or noisier input embeddings as the presentation time extends. This is further substantiated by the performance of the lower-performing backbones like DenseNet-121 and DenseNet-169. Although their absolute accuracy is low, they consistently gain accuracy with longer presentation time, $T_c$. This shows that given sufficient time, the SNN can better integrate and interpret even poor-quality spike trains embeddings to improve its final prediction. This finding is particularly notable, for example, in the case of the ResNet-50 model on the Chicken dataset, which registers a substantial accuracy improvement of over 20\% compared to the Phase 1 results, underscoring the unsupervised STDP classifier capacity to refine its output over time.

Finally, Tables \ref{tab:performance_chicken} and \ref{tab:performance_rabbit} present a summary of the achieved results and a comparison with the baseline models together with other state of the art models, when available. Its performance in terms of classification accuracy and energy efficiency are also shown. The results reveal a compelling dual advantage of the framework: it not only surpasses the performance of the original ANN backbones but also establishes a new state-of-the-art benchmark while operating with exceptional energy efficiency.

\begin{table*}[h!]
	\centering
	\caption{Performance comparison on the Chicken Dataset. The table contrasts the accuracy of the fine-tuned ANN backbones with our proposed Hybrid SNN models (at $T_c=300$ms). It also details the energy consumption for both model types, with the final column quantifying the energy efficiency improvement of the SNN. The last three rows list current state-of-the-art (SOTA) models for benchmarking purposes. Best results are highlighted in bold.}
	\label{tab:performance_chicken}
	\resizebox{\linewidth}{!}{
		\begin{tabular}{lrrrrr}
			\hline
			Backbone & ANN Accuracy (\%) &  SNN Accuracy (\%) & ANN Energy (J) & SNN Energy (J) & Improvement \\
			\hline
			resnet50       & 83,36              & 90,59                           & 5,17E-02             & 1,33E-05           & 3.895,43         \\
			resnet34       & 87,01              & 97,42                           & 4,60E-02             & 6,68E-06           & \textbf{6.878,45}         \\
			resnet18       & 84,21              & 95,63                           & 2,28E-02             & \textbf{5,00E-06}          & 4.555,80         \\
			mobilenet\_v2  & 86,96              & 94,51                           & \textbf{4,09E-03}             & 1,61E-05           & 253,43           \\
			mnasnet1\_0    & 86,58              & \textbf{98,32}                           & 4,20E-03             & 1,86E-05           & 225,75           \\
			densenet169    & \textbf{90,67}              & 60,92                           & 4,29E-02             & 1,04E-05           & 4.127,91         \\
			densenet121    & 90,54              & 69,43                           & 3,62E-02             & 1,11E-05           & 3.254,57         \\
			\hline
			\multicolumn{6}{l}{\textbf{SOTA models}} \\
			Xception \cite{kucukkara2025classification} & 96,40  & - & - & - & - \\
			ResTFG \cite{he2023reliable} & 96,90  & - & - & - & - \\
			ViT Hybrid SNN \cite{vazquez2024combining} & 87,60  & - & - & - & - \\
			Prunned EfficientNet-B0 \cite{acmali2024green} & 95,10  & - & - & - & - \\
			\hline
		\end{tabular}
	}
\end{table*}

\begin{table*}[h!]
	\centering
	\caption{Performance comparison on the Rabbit Dataset. The table contrasts the accuracy of the fine-tuned ANN backbones with our proposed Hybrid SNN models (at $T_c=300$ms). It also details the energy consumption for both model types, with the final column quantifying the energy efficiency improvement of the SNN. The last three rows list current state-of-the-art (SOTA) models for benchmarking purposes. Best results are highlighted in bold.}
	\label{tab:performance_rabbit}
	\resizebox{\linewidth}{!}{
		\begin{tabular}{lrrrrr}
			\hline
			Backbone & ANN Accuracy (\%) &  SNN Accuracy (\%) & ANN Energy (J) & SNN Energy (J) & Improvement \\
			\hline
			resnet50       & 82,67              & 91,05                           & 5,17E-02             & 1,14E-05           & 4.522,08         \\
			resnet34       & 82,87              & 96,91                           & 4,60E-02             & 6,55E-06           & \textbf{7.017,92}         \\
			resnet18       & 82,16              & 96,45                           & 2,28E-02             & \textbf{4,75E-06}           & 4.795,74         \\
			mobilenet\_v2  & 86,29              & 95,22                           & \textbf{4,09E-03}             & 1,43E-05           & 284,86           \\
			mnasnet1\_0    & 86,41              & \textbf{97,99}                           & 4,20E-03             & 1,82E-05           & 230,52           \\
			densenet169    & \textbf{90,96}              & 34,72                           & 4,29E-02             & 1,19E-05          & 3.611,85         \\
			densenet121    & 87,54              & 65,43                           & 3,62E-02             & 1,05E-05           & 3.449,94         \\
			\hline
			\multicolumn{6}{l}{\textbf{SOTA models}} \\
			ViT Hybrid SNN \cite{vazquez2024combining} & 81,50  & - & - & - & - \\ 
			Prunned EfficientNet-B0 \cite{acmali2024green} & 97,40  & - & - & - & - \\
			\hline
		\end{tabular}
	}
\end{table*}

Perhaps the most significant finding, observed consistently across both datasets, is that the conversion to a spiking neural network following the proposed procedure leads to a notable enhancement in classification accuracy for most architectures. This trend was exemplified by the MNASNet-1.0 model, which saw its accuracy climb from its ANN baseline of 86.51\% to a peak of 98.32\% on the Chicken dataset and 97.99\% on the Rabbit dataset, marking a new state of the art performance on this topic. Similarly, the ResNet family of models demonstrated substantial gains on both tasks, with ResNet-34?s accuracy increasing by over 10\% and 14\%, respectively. This consistent improvement suggests that the unsupervised spiking classifier effectively refines the feature representations provided by the converted backbone, regardless of the specific data domain. Concurrently, the architectural incompatibility of DenseNet models was also confirmed across both datasets, where they consistently exhibited a marked performance degradation post-conversion, mainly due to its excessive depth. Along this work it was observed that deeper models degrade the most as spiking information gets diluted in the network.

In addition, this enhancement in performance is invariably coupled with a significant reduction in energy consumption. This benefit was found to be universal across all models and datasets, confirming the foundational advantage of the SNN paradigm. The efficiency gains were profound, with improvement factors reaching remarkable levels. For instance, the ResNet-34 model required over 6,800 times less energy than its ANN counterpart on the Chicken dataset and over 7,000 times less on the Rabbit dataset. Even the highest-performing MNASNet-1.0 model became over 225 times more efficient on both datasets approximately. 

To further dive into the mechanisms behind the exceptional energy efficiency of the proposed framework, a deeper analysis of the classifier's internal dynamics is necessary. The model's efficiency is a direct consequence of its highly sparse and specialized neuronal activity. This principle is best illustrated by examining the behavior of the top-performing MNASNet model.

\begin{figure*}[!hbtp]
	\centering
	\begin{subfigure}[b]{0.4\textwidth}
		\centering 
		\includegraphics[width=\linewidth]{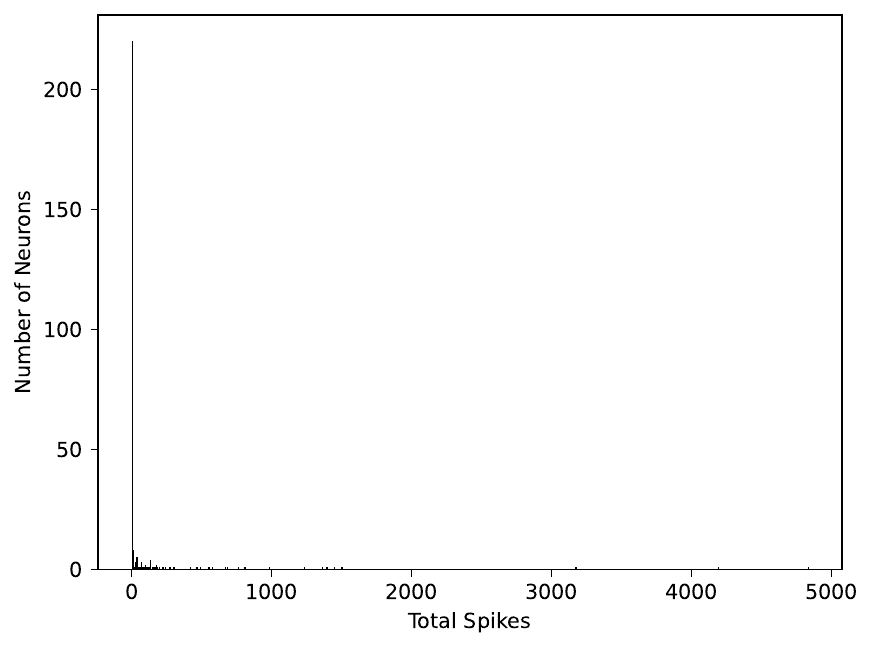}
		\caption{Chicken}.
	\end{subfigure}%
	~
	\begin{subfigure}[b]{0.4\textwidth}
		\centering 
		\includegraphics[width=\linewidth]{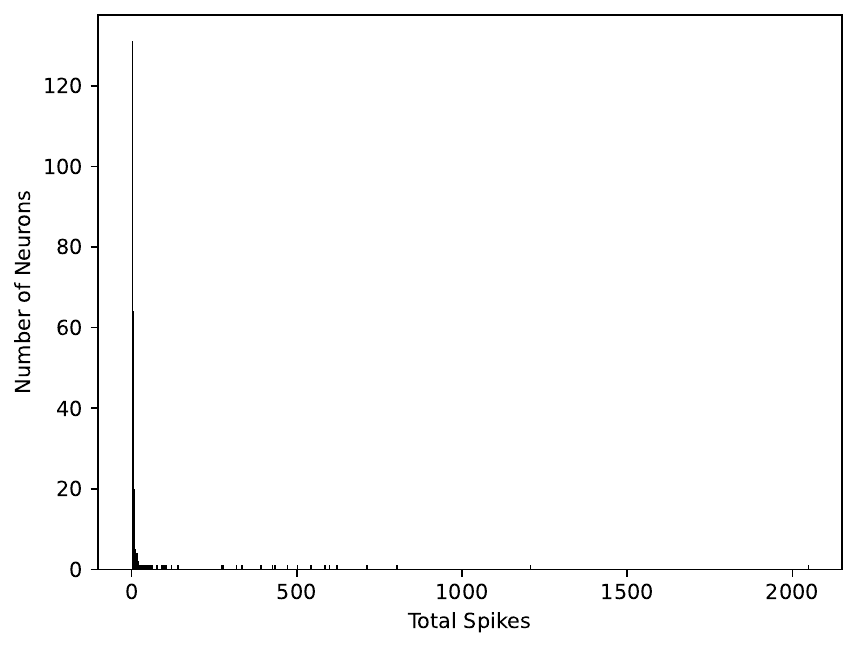}
		\caption{Rabbit}.
	\end{subfigure}
	\caption{Spike activity histograms for the MNASNet-1.0 STDP-based classifier on both datasets.}
	\label{fig:histograms}
\end{figure*}

The foundation of this efficiency is the sparse firing patterns, which are quantitatively demonstrated in the spike activity histogram in Fig \ref{fig:histograms}. The distribution in both datasets is heavily skewed, with a prominent peak near zero, indicating that the vast majority of neurons in the classifier fire a minimal number of spikes or remain entirely silent during inference. This extremely low computational load, where only a fraction of neurons participate in processing any given input, is the primary driver of the model's minuscule energy consumption.

\begin{figure*}[!hbtp]
	\centering
	\begin{subfigure}[b]{0.4\textwidth}
		\centering 
		\includegraphics[width=\linewidth]{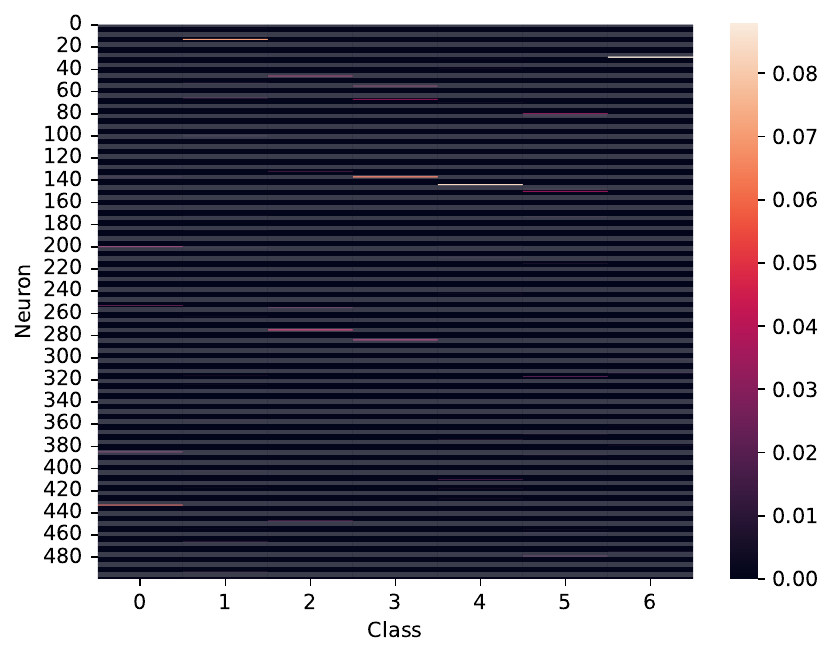}
		\caption{Neuron activity map}.
	\end{subfigure}%
	~
	\begin{subfigure}[b]{0.4\textwidth}
		\centering 
		\includegraphics[width=\linewidth]{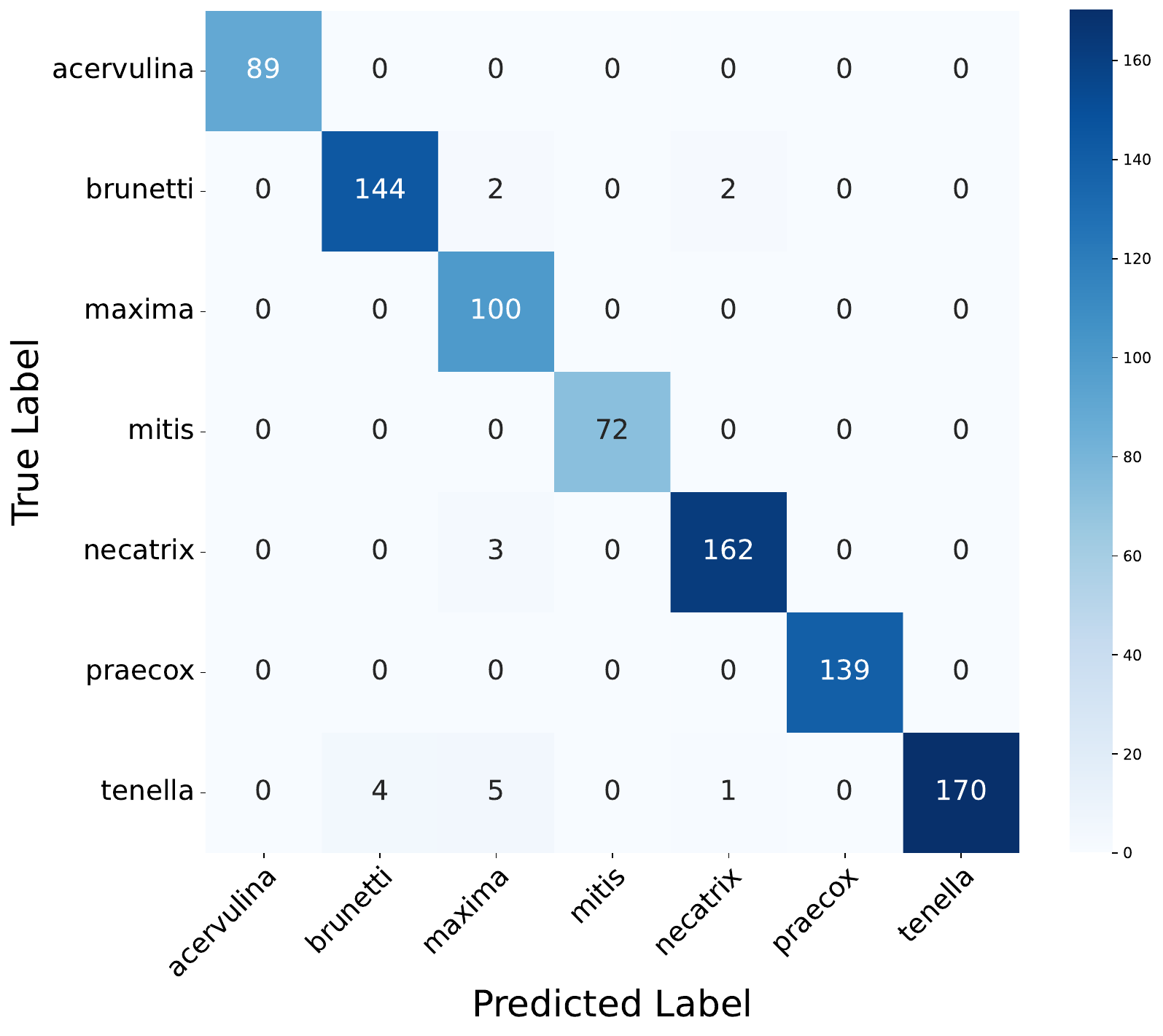}
		\caption{Confusion matrix}.
	\end{subfigure}
	\caption{Neuron activity map and confusion matrix for the chicken dataset.}
	\label{fig:chicken_map}
\end{figure*}

\begin{figure*}[!hbtp]
	\centering
	\begin{subfigure}[b]{0.4\textwidth}
		\centering 
		\includegraphics[width=\linewidth]{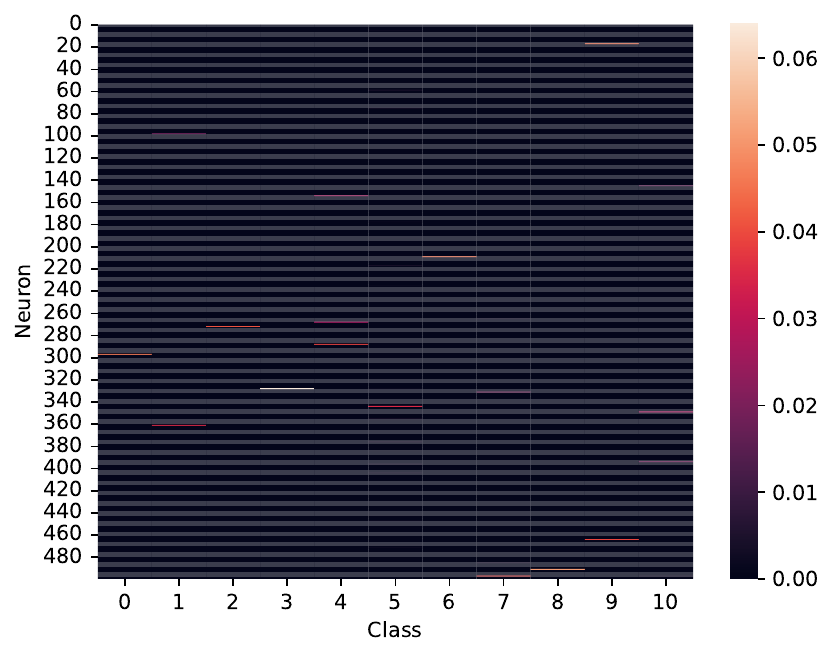}
		\caption{Neuron activity map}.
	\end{subfigure}%
	~
	\begin{subfigure}[b]{0.4\textwidth}
		\centering 
		\includegraphics[width=\linewidth]{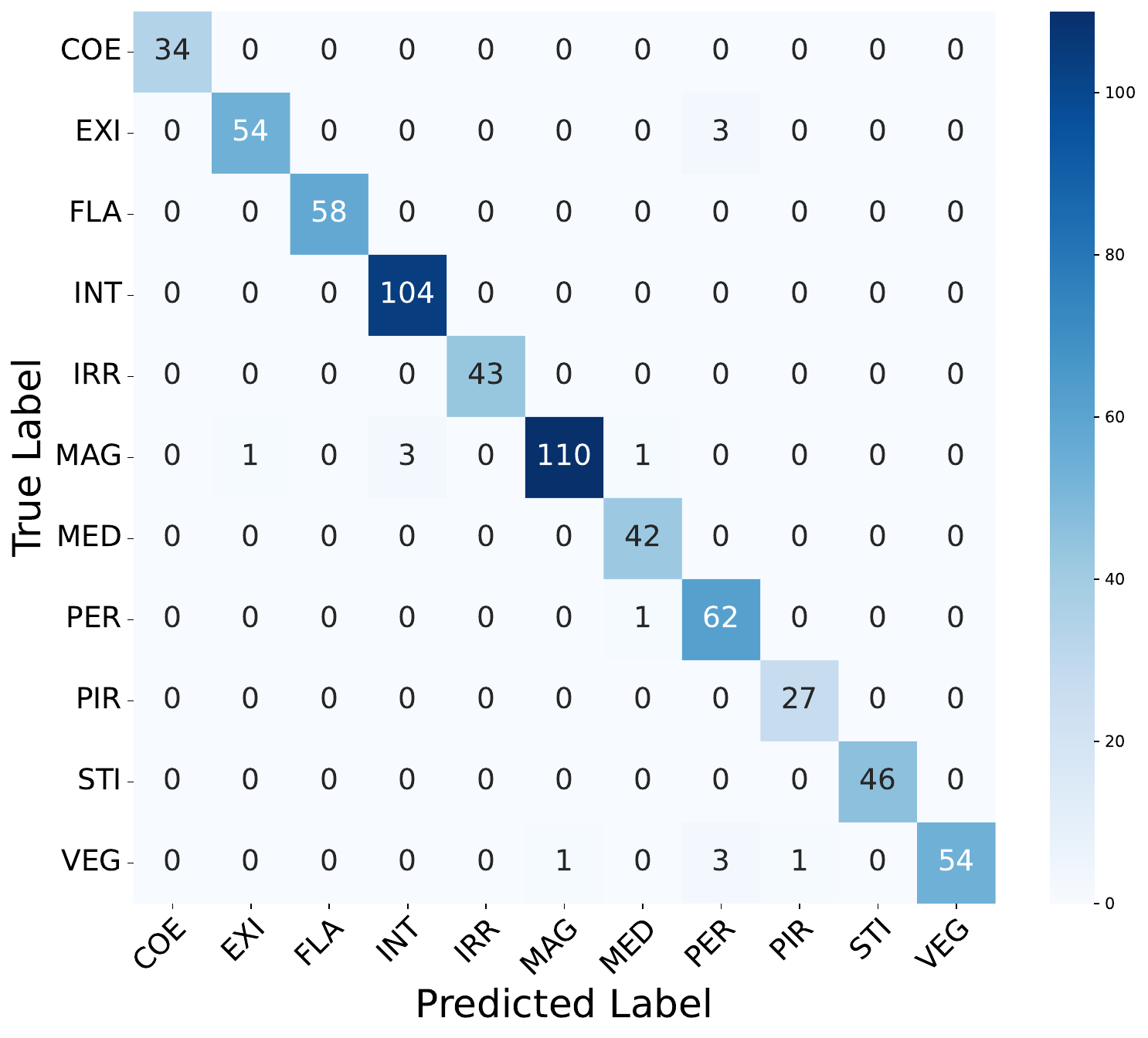}
		\caption{Confusion matrix}.
	\end{subfigure}
	\caption{Neuron activity map and confusion matrix for the rabbit dataset.}
	\label{fig:rabbit_map}
\end{figure*}

The mechanism enabling such profound sparsity is revealed in the neuron-to-class activity map, shown in Figs. \ref{fig:chicken_map} and \ref{fig:rabbit_map}, which provides critical insight into the classifier's learned representation. This map shows the average firing rate for each class, so it allows us to see which neurons are more active to a given class. In both cases, the map in general is overwhelmingly dark, signifying that for each class, only a small and distinct ensemble of neurons becomes highly active. This demonstrates a powerful form of neuronal specialization of the lateral inhibition process, where the network has learned to assign a unique and compact group of neurons to each specific category. For example, the neurons responsible for identifying Class 3 in chicken data (mitis) show strong activation, while remaining almost completely dormant for all other classes. This learned specialization ensures that the network's response to any input is localized to a tiny subset of its total neuronal population, thereby maintaining overall sparsity and efficiency.

The consequence of this highly specialized and sparse coding scheme is not a compromise in performance but rather an enhancement of it. The confusion matrices shows a strongly diagonal structure, indicative of high classification accuracy and excellent discriminative power. This high performance is a direct result of the clear separation between neuronal ensembles seen in the activity map. Because the neuronal groups assigned to different classes have minimal overlap, the model makes confident and accurate predictions with very few misclassifications.

However, despite the high accuracy and energy efficiency of the proposed hybrid architecture, certain limitations must be acknowledged. The most significant of these is the inherent trade-off between energy consumption and inference latency. The sequential, timestep-based processing of SNNs introduces a latency that scales with the presentation times $T_b$ and $T_c$. However, this trade-off was a deliberate design choice, guided by the specific requirements of the target application: providing a diagnostic tool for livestock on remote farms where access to energy is limited. In this context, a high latency acceptable if energy consumption is significantly reduced. As the primary objective was the minimization of energy, inference speed was permissibly sacrificed.


%
%
\section{Conclusions} \label{sec:conclusions}

This work has addressed the need for an accurate and energy-efficient diagnostic tool for Eimeria parasites, a significant threat to the poultry and rabbit industries. While conventional deep learning models offer high accuracy, their computational expense limits their deployment in resource-constrained field environments. To overcome this barrier, this paper introduced a novel approach for transfer learning in spiking neural network that successfully integrates the feature extraction power of converted ANNs with the profound efficiency of an unsupervised, STDP-based spiking classifier.

The primary contribution of this work is a two-stage methodology that establishes a new paradigm for transfer learning in SNNs. Through this approach, we have demonstrated that ANN-to-SNN conversion does not have to result in a performance trade-off. On the contrary, the proposed methodology enhances classification accuracy beyond their ANN as demonstrated by empirical results The proposed framework achieved a new state-of-the-art accuracy of 98.32\% with the MNASNet-1.0 backbone, with an extraordinary reduction in computational energy comsumption. The models achieved significant energy savings, often by several orders of magnitude, across all tested architectures. For example, the ResNet-34 model became over 6,800 times more efficient. Similarly, the best-performing model was 225 times more efficient than its ANN counterpart.

The mechanism behind this dual success was shown to be the network's ability to learn a sparse and highly specialized neural code, where distinct neuronal ensembles are selectively activated for each parasite class. This ensures that state-of-the-art performance is achieved with minimal computational overhead, aligning perfectly with the principles of Green AI. By leveraging an unsupervised learning mechanism, our architecture paves the way for autonomous, on-chip systems that can be deployed on low-power neuromorphic hardware.

\section*{Acknowledgements}

This work was supported by the Spanish Ministry of Science and Innovation with project PID2023-149511OB-I00 and under the programme for mobility stays at foreign higher education and research institutions "José Castillejo Junior" with code CAS23/00340.

\input{sn-article.bbl}

\appendix

\section{Full results tables}

\begin{table*}[hbtp!]
	\centering
	\caption{Classification accuracy (\%) on the chicken dataset for all backbone models across different simulation timesteps ($T_b$). The baseline performance in ANN-mode ($T_b=0$) is shown in the first column. }
	\label{tab:phase1_chicken_results}
	\resizebox{\textwidth}{!}{%
		\begin{tabular}{lr|rrrrrrr}
			\hline
			\textbf{Backbone Model} & \multicolumn{8}{c}{\textbf{Simulation Timesteps ($T_b$)}} \\ 
			\hline \hline                                                                              
			& \textbf{0 (ANN)} & \textbf{4} & \textbf{8} & \textbf{16} & \textbf{32} & \textbf{64} & \textbf{128} & \textbf{256} \\ \hline
			ResNet-18 & 95,62  & 32,29  & 56,16  & 77,16  & 90,25  & 94,14  & 94,88  & \textbf{95,13}  \\ 
			ResNet-34 & 98,47  & 23,31  & 33,51  & 50,14  & 74,82  & 92,53  & 96,42  & \textbf{97,40}  \\ 
			ResNet-50 & 97,27  & 21,74  & 18,98  & 18,37  & 23,66  & 45,77  & 72,27  & \textbf{90,83}  \\ 
			ResNet-101 & 97,68  & 18,11  & 16,89  & 15,37  & 22,46  & 39,93  & 55,91  & \textbf{64,68}  \\ 
			ResNet-152 & 97,11  & 19,20  & 19,27  & 30,23  & 55,10  & \textbf{70,89 } & 50,71  & 38,53  \\ 
			MNASNet-0.5 & 14,29  & 14,29  & 14,29  & 14,29  & 14,29  & 14,29  & 14,29  & 14,29  \\ 
			MNASNet-0.75 & 60,62  & 16,56  & 18,45  & 24,28  & 27,97  & 32,08  & 39,76  & \textbf{47,26}  \\ 
			MNASNet-1.0 & 98,41  & 15,72  & 20,97  & 51,33  & 91,13  & 96,13  & 96,90  & \textbf{97,37}  \\ 
			MNASNet-1.3 & 83,61  & 30,39  & 40,07  & 52,05  & 66,80  & 78,81  & 81,37  & \textbf{81,70}  \\ 
			VGG-11 & 14,29  & 14,29  & 14,29  & 14,29  & 14,29  & 14,29  & 14,29  & 14,29  \\ 
			VGG-16 & 14,29  & 14,29  & 14,29  & 14,29  & 14,29  & 14,29  & 14,29  & 14,29  \\ 
			EfficientNetV2-S & 95,85  & 14,29  & 14,29  & 14,29  & 14,29  & 14,29  & 14,29  & 14,29  \\ 
			EfficientNetV2-M & 86,19  & 14,29  & 14,29  & 14,29  & 14,29  & 14,29  & 14,08  & 14,29  \\ 
			EfficientNet-B0 & 96,44  & 14,31  & 16,91  & 15,31  & 16,95  & 21,09  & 36,16  & \textbf{64,38}  \\ 
			EfficientNet-B1 & 96,46  & 15,45  & 14,44  & 14,29  & 14,75  & 18,17  & \textbf{23,91}  & 17,82  \\ 
			EfficientNet-B2 & 94,00  & 14,29  & 14,29  & 14,29  & 14,29  & 14,29  & 14,29  & 14,29  \\ 
			EfficientNet-B3 & 96,65  & 14,29  & 14,29  & 19,47  & 34,28  & 41,15  & 56,39  & \textbf{73,53}  \\ 
			EfficientNet-B4 & 85,54  & 14,29  & 16,16  & 14,52  & 13,52  & 14,12  & 13,75  & \textbf{24,31}  \\ 
			EfficientNet-B5 & 96,99  & 14,29  & 14,29  & 35,11  & 40,79  & 48,90  & 66,45  & \textbf{82,04}  \\ 
			MobileNetV2 & 98,35  & 21,01  & 15,62  & 18,30  & 33,08  & 60,18  & 83,05  & \textbf{93,15}  \\ 
			MobileNetV3-Small & 95,87  & 14,42  & 14,41  & 14,11  & 14,30  & 19,76  & 17,92  & \textbf{21,25}  \\ 
			MobileNetV3-Large & 97,47  & 15,47  & 15,88  & 15,71  & 18,85  & 35,56  & 74,05  & \textbf{92,96}  \\ 
			DenseNet-121 & 98,05  & 17,01  & 22,05  & 38,32  & 58,48  & 73,06  & 87,04  & \textbf{94,27}  \\ 
			DenseNet-169 & 97,45  & 24,31  & 33,07  & 57,39  & 76,49  & 87,70  & \textbf{91,45}  & 90,35  \\ 
			ViT-B/16 & 14,29  & 14,29  & 14,29  & 14,29  & 14,29  & 14,29  & 14,29  & 14,29  \\ \hline
		\end{tabular}
	}
\end{table*}

\begin{table*}[hbtp!]
	\centering
	\caption{Classification accuracy (\%) on the rabbit dataset for all backbone models across different simulation timesteps ($T_b$). The baseline performance in ANN-mode ($T_b=0$) is shown in the first column.}
	\label{tab:phase1_rabbit_results}
	\resizebox{\textwidth}{!}{%
		\begin{tabular}{lr|rrrrrrr}
			\hline
			\textbf{Backbone Model} & \multicolumn{8}{c}{\textbf{Simulation Timesteps ($T_b$)}} \\ 
			\hline \hline                                                                               
			& \textbf{0 (ANN)} & \textbf{4} & \textbf{8} & \textbf{16} & \textbf{32} & \textbf{64} & \textbf{128} & \textbf{256} \\ \hline
			ResNet-18 & 98,38  & 37,30  & 62,54  & 87,14  & 95,99  & 97,63  & 98,26  & \textbf{98,33}  \\ 
			ResNet-34 & 97,54  & 34,73  & 57,75  & 75,86  & 87,86  & 95,64  & 97,42  & \textbf{97,46 } \\ 
			ResNet-50 & 98,36  & 28,54  & 28,37  & 36,88  & 56,62  & 78,54  & 92,90  & \textbf{94,99}  \\ 
			ResNet-101 & 98,08  & 24,64  & 30,36  & 45,18  & \textbf{57,23}  & 53,49  & 41,59  & 33,62  \\ 
			ResNet-152 & 97,62  & 34,96  & 59,20  & 73,84  & \textbf{77,87}  & 65,42  & 50,08  & 38,81  \\ 
			MNASNet-0.5 & 9,09  & 9,09  & 8,88  & 9,09  & 8,59  & 9,09  & 9,16  & 9,09  \\ 
			MNASNet-0.75 & 9,09  & 9,09  & 9,09  & 9,09  & 9,09  & 9,09  & 9,09  & 9,09  \\ 
			MNASNet-1.0 & 98,01  & 11,67  & 14,55  & 45,57  & 80,94  & 95,20  & \textbf{98,06}  & \textbf{98,06}  \\ 
			MNASNet-1.3 & 9,09  & 9,09  & 9,09  & 9,09  & 9,09  & 9,09  & 9,09  & 9,09  \\ 
			VGG-11 & 9,09  & 9,09  & 9,09  & 9,09  & 9,09  & 9,09  & 9,09  & 9,09  \\ 
			VGG-16 & 9,09  & 9,09  & 9,09  & 9,09  & 9,09  & 9,09  & 9,09  & 9,09  \\ 
			EfficientNetV2-S & 94,75  & 9,09  & 9,09  & 9,24  & 11,62  & 13,34  & 12,73  & \textbf{14,23}  \\ 
			EfficientNetV2-M & 25,81  & 9,09  & 9,09  & 9,09  & 9,09  & 9,09  & 8,77  & 9,37  \\ 
			EfficientNet-B0 & 94,68  & 11,94  & 16,05  & 24,83  & 34,93  & 35,94  & 34,91  & \textbf{37,88}  \\ 
			EfficientNet-B1 & 96,81  & 9,09  & 8,19  & 7,74  & 7,84  & 9,09  & 9,09  & 12,98  \\ 
			EfficientNet-B2 & 94,31  & 9,09  & 9,09  & 9,09  & 9,09  & 9,09  & 9,09  & 9,09  \\ 
			EfficientNet-B3 & 97,20  & 14,90  & 16,32  & 28,87  & 51,21  & 63,45  & 71,79  & \textbf{79,16}  \\ 
			EfficientNet-B4 & 80,78  & 9,09  & 9,09  & 9,25  & 12,79  & 11,39  & 10,74  & 15,30  \\ 
			MobileNetV2 & 97,64  & 29,00  & 41,93  & 53,86  & 74,45  & 94,67  & \textbf{97,75}  & 96,39  \\ 
			MobileNetV3-Small & 95,23  & 9,09  & 9,09  & 9,09  & 9,09  & 9,09  & 9,09  & 9,09  \\ 
			MobileNetV3-Large & 97,89  & 9,25  & 9,11  & 9,71  & 19,43  & 28,45  & 70,50  & \textbf{86,77}  \\ 
			DenseNet-121 & 98,63  & 14,17  & 18,16  & 25,55  & 42,98  & 64,10  & 81,86  & \textbf{87,10}  \\ 
			DenseNet-169 & 98,22  & 13,18  & 47,20  & 63,04  & \textbf{72,36}  & 63,75  & 48,89  & 30,75  \\ 
			ViT-B/16 & 9,09  & 9,09  & 9,09  & 9,09  & 9,09  & 9,09  & 9,09  & 9,09  \\ \hline
		\end{tabular}
	}
\end{table*}


\begin{table*}[!ht]
	\centering
	\caption{Accuracy (\%) of the different backbones using the unsupervised STDP classification layer for several presentation times. Results for the chicken dataset.}
	\label{tab:SNN_accuracy_chicken}
	\begin{tabular}{lrrrrr}
		\hline
		& \multicolumn{5}{c}{Presentation Time (ms)} \\
		\hline
		Backbone & 100 & 150 & 200 & 250 & 300 \\ \hline \hline
		resnet50 & 82,30 & 87,46 & 90,03 & 89,24 & \textbf{90,59} \\ 
		resnet34 & 95,74 & 96,19 & 95,85 & 96,52 & \textbf{97,42} \\ 
		resnet18 & 95,40 & \textbf{96,08} & 95,96 & 96,08 & 95,63 \\ 
		mobilenet\_v2 & 88,57 & 91,15 & 93,28 & 94,17 & \textbf{94,51} \\ 
		mnasnet1\_0 & 97,53 & 97,98 & 98,09 & 98,09 & \textbf{98,32} \\ 
		densenet169 & 42,44 & 46,80 & 49,27 & 54,64 & \textbf{60,91} \\ 
		densenet121 & 52,51 & 54,31 & 60,69 & 62,37 & \textbf{69,42} \\ \hline
	\end{tabular}
\end{table*}

\begin{table*}[!ht]
	\centering
	\caption{Accuracy (\%) of the different backbones using the unsupervised STDP classification layer for several presentation times. Results for the rabbit dataset.}
	\label{tab:SNN_accuracy_rabbit}
	\begin{tabular}{lrrrrr}
		\hline
		& \multicolumn{5}{c}{Presentation Time (ms)} \\
		\hline
		Backbone & 100 & 150 & 200 & 250 & 300 \\ \hline \hline
		resnet50 & 86,11 & 88,73 & 88,73 & 89,66 & \textbf{91,04} \\ 
		resnet34 & 90,43 & 94,59 & 94,90 & 95,37 & \textbf{96,91} \\ 
		resnet18 & 92,59 & 94,75 & 94,44 & 95,67 & \textbf{96,45} \\ 
		mobilenet\_v2 & 91,66 & 92,90 & 92,59 & 93,20 & \textbf{95,21} \\ 
		mnasnet1\_0 & 94,13 & 94,90 & 94,90 & 95,52 & \textbf{97,99} \\ 
		densenet169 & 28,10 & 29,16 & 30,55 & 31,17 & \textbf{34,72} \\ 
		densenet121 & 52,77 & 54,93 & \textbf{67,28} & 61,41 & 65,43 \\ \hline
	\end{tabular}
\end{table*}

\end{document}